\documentclass[times, 5p]{elsarticle}
\usepackage{hyperref}
\journal{-}

\usepackage{adjustbox}
\usepackage{lineno}
\usepackage{amsmath}
\usepackage{diagbox}
\usepackage{subfigure}
\usepackage{listings} %

\usepackage[utf8]{inputenc}
\usepackage{booktabs}
\usepackage{multirow}
\usepackage{geometry}
\usepackage{tabularx}

\DeclareFixedFont{\ttb}{T1}{txtt}{bx}{n}{8} %
\DeclareFixedFont{\ttm}{T1}{txtt}{m}{n}{8}  %

\usepackage{color}
\definecolor{deepblue}{rgb}{0,0,0.5}
\definecolor{deepred}{rgb}{0.6,0,0}
\definecolor{deepgreen}{rgb}{0,0.5,0}

\definecolor{dkgreen}{rgb}{0,0.6,0}
\definecolor{mauve}{rgb}{0.58,0,0.82}
\definecolor{desertpurple}{rgb}{0.525490,0.176470,0.525490}

\newcommand\pythonstyle{\lstset{
language=Python,
basicstyle=\ttm,
numberstyle={\tiny},
otherkeywords={self,as},             %
keywordstyle=\ttb\color{deepblue},
commentstyle=\color{mauve},
emph={MyClass,__init__},          %
emphstyle=\ttb\color{deepred},    %
stringstyle=\color{deepgreen},
frame=tb,                         %
showstringspaces=false            %
}}

\lstnewenvironment{python}[1][]
{
\pythonstyle
\lstset{#1}
}
{}

\usepackage{comment}
\newsavebox{\measurebox}

\newcommand{\lossT}{\mathcal{L}_{\mathcal{T}}}

\hypersetup{
    colorlinks,
    linkcolor={red!50!black},
    citecolor={blue!50!black},
    urlcolor={blue!50!black}
}

\def \tnorm {\top}
\def \conorm {\bot}

\biboptions{authoryear}

\begin{document}

\begin{frontmatter}
\title{SSG2: A new modelling paradigm for semantic segmentation}
\author[data61]{Foivos I. Diakogiannis\fnref{myfootnote1}}
\author[data61]{Suzanne Furby}
\author[data61]{Peter Caccetta}
\author[data61]{Xiaoliang Wu}
\author[uniastra]{Rodrigo Ibata}
\author[imt]{Ondrej Hlinka}
\author[data61]{John Taylor}

\address[data61]{Data61, CSIRO, Floreat WA, Australia}
\address[uniastra]{University of Strasbourg, France}
\address[imt]{IM\&T CSIRO,  Australia}
\fntext[myfootnote1]{foivos.diakogiannis@data61.csiro.au}

\begin{abstract}
State-of-the-art models in semantic segmentation primarily operate on single, static images, generating corresponding segmentation masks. This one-shot approach leaves little room for error correction, as the models lack the capability to integrate multiple observations for enhanced accuracy.
Inspired by work on semantic change detection, we address this limitation by introducing a methodology that leverages a sequence of observables generated for each static input image. By adding this ``temporal'' dimension, we exploit strong signal correlations between successive observations in the sequence to reduce error rates.
Our framework, dubbed SSG2 (Semantic Segmentation Generation 2), employs a dual-encoder, single-decoder base network augmented with a sequence model. The base model learns to predict the set intersection, union, and difference of labels from dual-input images. 
Given a fixed target input image and a set of support images, the sequence model builds  the predicted mask of the target by synthesizing the partial views from each sequence step and filtering out noise.  %
We evaluate SSG2 across three diverse datasets: UrbanMonitor, featuring orthoimage tiles from Darwin, Australia with five spectral bands and 0.2m spatial resolution; ISPRS Potsdam, which includes true orthophoto images with multiple spectral bands and a 5cm ground sampling distance; and ISIC2018, a medical dataset focused on skin lesion segmentation, particularly melanoma. The SSG2 model demonstrates rapid convergence within the first few tens of epochs and significantly outperforms UNet-like baseline models with the same number of gradient updates. %
However, the addition of the temporal dimension results in an increased memory footprint. While this could be a limitation, it is offset by the advent of higher-memory GPUs and coding optimizations.
\end{abstract}


\end{frontmatter}

\begin{figure}
\begin{center}
\includegraphics[clip, trim=0.25cm 0.0cm 0.0cm 
0.1cm,width=\columnwidth]{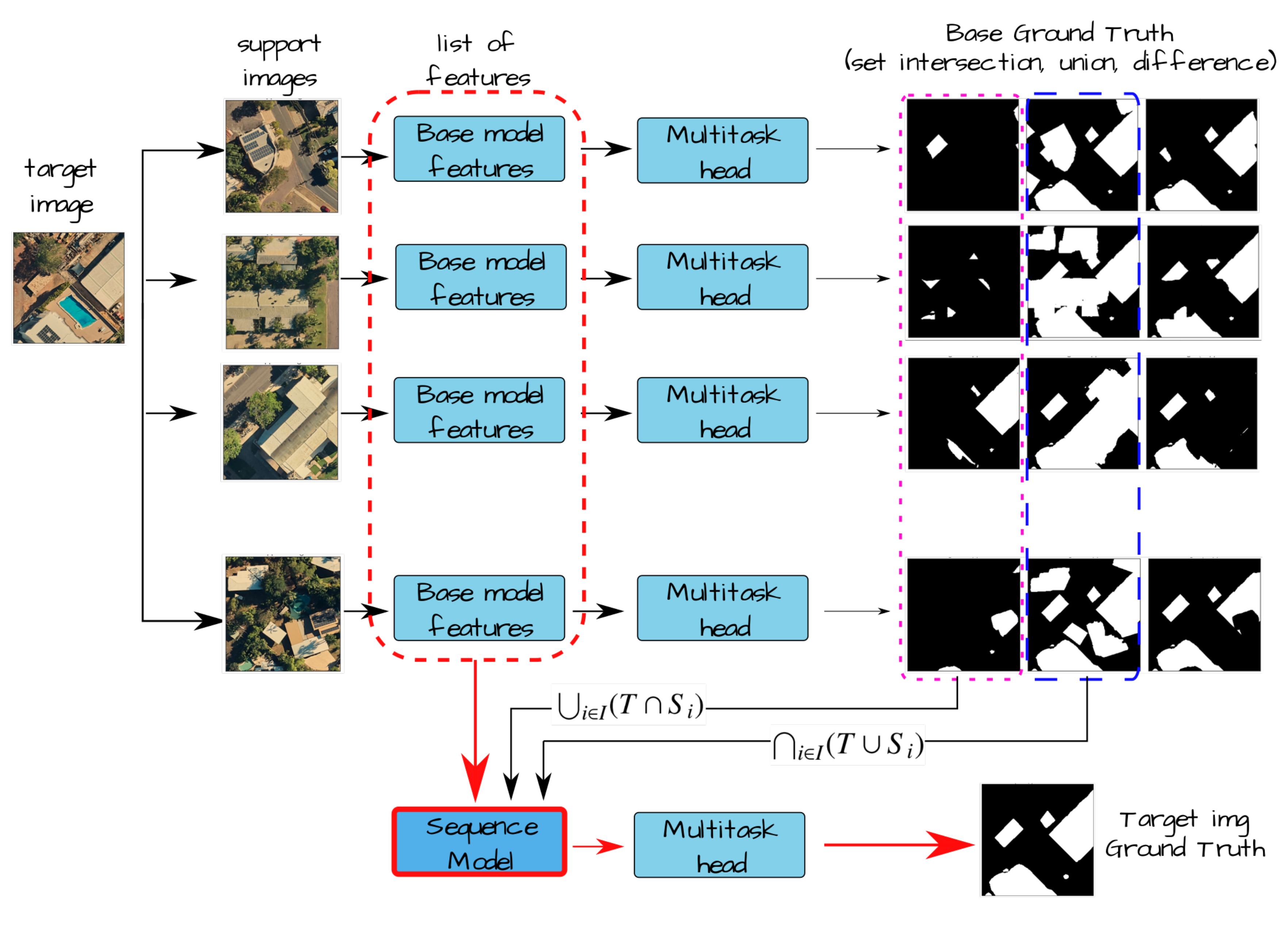}
\end{center}
\caption{Overview of the SSG2 modelling framework. A central tenet of this design is that the union of all intersecting ground truths, as well as the intersection of all unions, serves as a strong prior to approximate the Target image's actual ground truth. Crucially, true feature signals across sequence elements are strongly correlated, contrasting with uncorrelated noise (prediction errors), thereby boosting the algorithm's performance.}
\label{system_sequence_model}
\end{figure}

\begin{figure*}[!h]
\begin{center}
\includegraphics[clip, trim=1.25cm 4.0cm 0.25cm 
3.1cm,width=\textwidth]{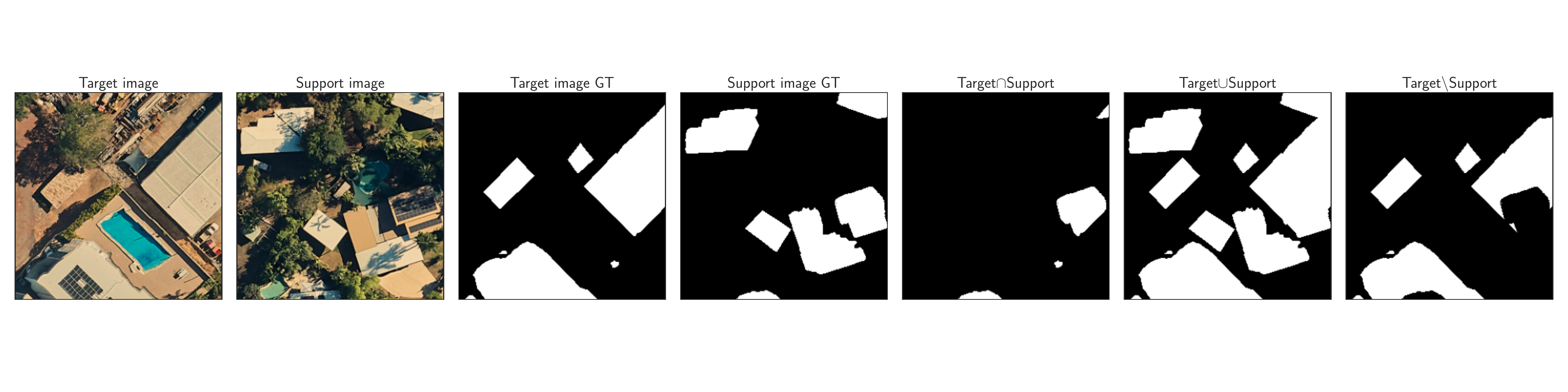}
\end{center}
\caption{Example of input images and ground truth to the base model. The base model uses as ground truth the set intersection, union and difference between the Target and Support image. Note that the difference operation breaks the symmetry of the two inputs. }
\label{base_model_inputs_n_masks}
\end{figure*}

\section{ Introduction }

Semantic segmentation is a useful tool in various scientific applications, providing a more nuanced understanding of spatial data. In remote sensing, it helps with tasks like land cover classification and natural resource monitoring, turning complex imagery into more digestible information \citep{8113128}. Similarly, in medical imaging, it aids in identifying specific tissues and potential anomalies, thereby assisting in diagnosis and treatment plans \citep{LITJENS201760} . 

In the existing landscape of semantic segmentation, deep learning models are primarily structured around an image-to-image framework \citep{taghanaki2019deep,MO2022626}, mostly dominated by the very successful UNet-like architectures \citep{9446143}. These models, while effective, are constrained to single-image inputs for pixel-wise annotation, thereby lacking a mechanism for incorporating multiple observations for potential statistical refinement. 
Ideally, we would employ multiple observations of the same input, taken at different times, to enable statistical averaging from model predictions and thereby increase the Signal-to-Noise Ratio (SNR). This approach is not new; in astronomy \citep{Kurczynski_2010}, and radio astronomy \citep{10.1093/rasti/rzad002}, image stacking techniques, ranging from simple averaging to advanced methods, work to emphasize consistent signals and nullify random noise\footnote{e.g. \href{DeepSkyStacker}{http://deepskystacker.free.fr/english/index.html}}. Similarly, in MRI, various techniques have been developed to improve image quality. For example, correlating two acquisitions of the same MR image can significantly improve the SNR and the Contrast-to-Noise Ratio (CNR) with minimal loss of resolution \citep{SIJBERS19961157}. Both fields demonstrate the practical benefits of utilizing multiple observations to enhance data quality. For deep learning problems though, this is not straightforward, given that we only have static imagery as inputs with corresponding annotated masks. 

So how can we best leverage existing annotated datasets to introduce the benefits of multiple observations into semantic segmentation? To address this, we transition to a sequence-of-images-to-image framework. In this revised framework, inspired from change detection models in deep learning, a sequence is constructed from a single target image of interest, paired with a set of support images. Each sequence element, therefore, comprises this target image and a corresponding support image, prompting the query: \emph{what and where are the similarities between these two images}? By comparing the target with the set of support images, the framework offers the potential for statistical filtering, rooted in the strong correlation of true signals and the uncorrelated nature of noise across different sequence elements. In set theoretic terms, this question translates to identifying the set intersection of the ground truth masks, attributing "same location, same class" to the identified regions. However, when dealing with small or sparsely represented objects, the set intersection frequently results in primarily empty space, leading to a dataset imbalanced towards negative classes. To rectify this, we augment the set operations to include set union and set difference, enriching the class distribution in the training dataset and aiding convergence.

In the present study, we introduce a novel framework for semantic segmentation, termed SSG2\footnote{Code release \href{https://github.com/feevos/ssg2}{https://github.com/feevos/ssg2}.} (Semantic Segmentation Generation 2), that incorporates a dual-input scheme, comprising a target image and a set of support images (Figures \ref{system_sequence_model} and \ref{base_model_inputs_n_masks}). Through pairwise comparisons between the target and each support image, the algorithm synthesizes a series of observables that not only expedite convergence but also substantially reduce error rates.  Concurrently, we unveil a specialized attention mechanism specifically engineered for integration with hybrid convolutional -- Transformer architectures in computer vision tasks. Additionally, the work features a new activation function designed to ameliorate the challenges associated with gradient explosion in deep neural networks.

In the subsequent sections, we commence by delineating the datasets employed for evaluating our methodology, followed by an in-depth exposition of the constituent elements of our deep learning architecture and modelling design.

\section{Methods}

\subsection{Data}

We evaluate our approach on two Very High Resolution remote sensing datasets, the primary focus of our work, as well as a medical imaging dataset.

The first dataset is the  \citep{isprs_bsf_swissphoto_2023}  Potsdam dataset. The dataset is comprised of a subset of true orthophoto (TOP) images taken from a larger mosaic, as well as a Digital Surface Model (DSM). The TOP features four spectral bands in the visible (VIS) range, including red (R), green (G), blue (B), and near-infrared (NIR), with a ground sampling distance of 5 cm. The normalized DSM layer offers height information for each pixel, as it has had the ground elevation removed. For training the semantic segmentation models, the four spectral bands (VISNIR) and the normalized DSM were combined (VISNIR + DSM). The annotations include six categories: impervious surfaces, buildings, cars, low vegetation, trees, and background.

\begin{figure}
\begin{center}
\includegraphics[width=\columnwidth]{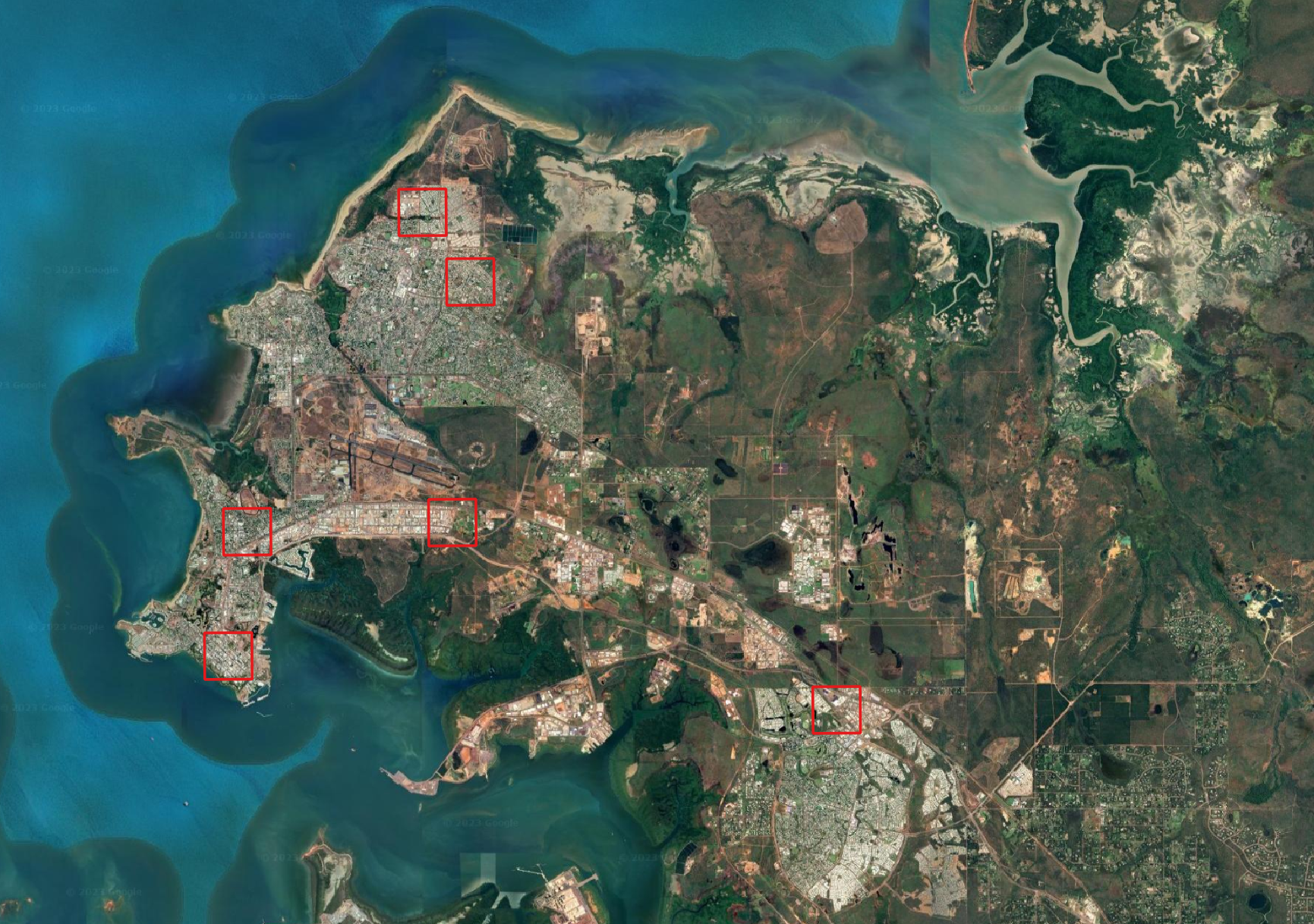}
\end{center}
\caption{Darwin dataset area selection. Background imagery  Map data \textcopyright 2023 Google}
\label{darwin_image}
\end{figure}

The second dataset, UrbanMonitor, offers a closer approximation to real-world conditions compared to more academically curated datasets like ISPRS Potsdam. Comprising 6 orthoimage tiles from various environments -- urban, commercial, residential, rural or mixed -- these tiles span locations across the Darwin region in Northern Territory, Australia (shown in red squares in Fig. \ref{darwin_image}). Each tile covers an area of 1.2 km by 1.2 km with a spatial resolution of 0.2m and includes five bands: red, green, blue, near-infrared, and a normalized DSM (NSM) band. Captured in August 2021 using a photogrammetric aerial camera system (PhaseOne camera), the DSM/NSM was generated from stereo images.

The ground truth masks for UrbanMonitor were primarily generated using automated semantic segmentation models, a pre-trained UNet-like architecture ResUNet-a  \citep{DIAKOGIANNIS202094}, which was developed in a previous study. These automated annotations were subsequently refined through some manual editing. This approach to ground truth labeling reflects real-world constraints, such as budget and time, making the dataset a practical testbed for evaluating the robustness of our proposed algorithms.

Motivated by the cross-disciplinary success of UNet-like architectures in semantic segmentation tasks, we extend our investigation to include ISIC 2018 \citep{DBLP:journals/corr/abs-1902-03368,tschandl2018ham10000}, a dataset distinct from our primary focus on remote sensing. Despite its comprehensive nature, this collection is relatively sparse, consisting of only 2,594 training images -- making it much smaller than other commonly used datasets like CIFAR10 or ImageNet. As part of the International Skin Imaging Collaboration project, ISIC 2018 serves as a benchmark in dermoscopic image analysis for skin cancer diagnosis. It comprises high-quality skin lesion images with expert annotations for various classes of skin diseases, including melanoma and non-melanoma types. The inclusion of this sparse dataset allows us to evaluate the algorithm's performance in a different scientific domain, while also providing a more comprehensive validation of its capabilities, particularly in contexts where data sparsity is a concern. This resource has also found widespread use in training and evaluating machine learning models, thus serving as a valuable asset in medical imaging research.

\subsection{Patch Tanimoto Attention: A ViT-like Attention for Convolutions}

\begin{figure*}[ht!]
\begin{center}
\includegraphics[width=\textwidth]{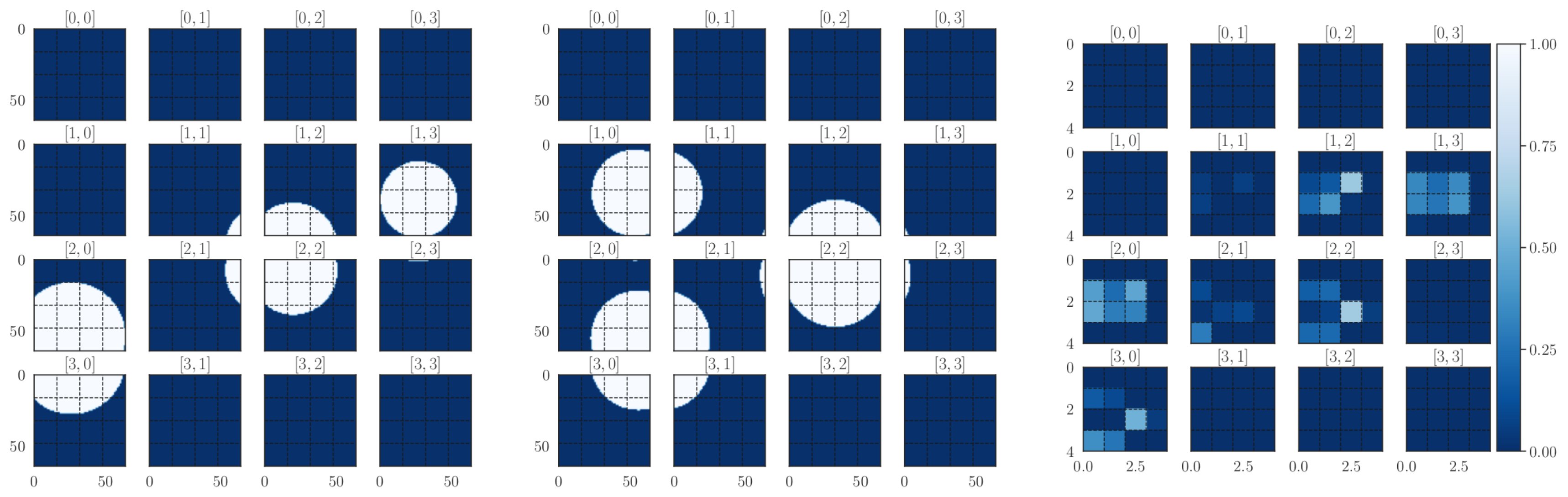}
\includegraphics[width=\textwidth]{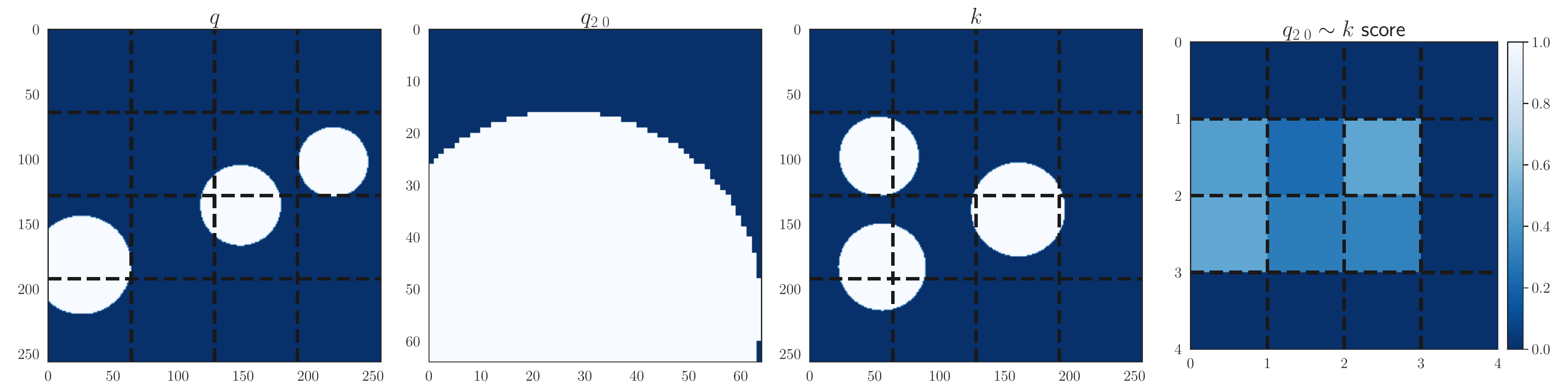}
\end{center}
\caption{Example of query, $q$, $k$ similarity for two $256\times 256$ binary images of disks. Top left panel: query features, top middle panel: key features. Top right panel: coordinate map that results from comparing query, $q$ with key, $k$ features. This map has all the information needed to identify cross spatial correlations of $q$. In the bottom row we show the similarity map (from left to right, fourth panel) that results from the comparison of the $q_{20}$ element of the query (second panel), with the key, $k$ (third panel).}
\label{patch_attention}
\end{figure*}

The Visual Transformer \citep[][hereafter ViT]{DBLP:journals/corr/abs-2010-11929} makes a significant contribution by introducing patch-based image processing \citep[][see also \citealt{weng2023transformer}]{trockman2023patches}. This technique allows for the comparison of patches across different spatial locations within an image, enabling correlation of information existing in distinct spatial areas. In contrast, standard convolutions are limited by their local nature and cannot perform such comparisons.

Inspired by this observation, we design an attention module, tailored for hybrid convolutional - transformer models, that keeps the ability to spatially correlate distinct areas of an input image. This comes without the need for an explicit positional embedding. 
The key to achieving this is to construct a correlation coordinate map, $q\sim k$  from the comparison of query, $q$,  and key, $k$, that preserves the spatial ordering of values $v$ and that encapsulates both spatial correlation and similarity information. This coordinate map - which by construction has position information encoded - is then used to select which entries of the values, $v$, tensor should be used for the particular query, $q$ image. Importantly, by proper reordering of the entries of the value $v$ tensor, the selection is done by element-wise multiplication, thus keeping the dimensionality of intermediate tensors as small as possible. That is, comparison does not take place with matrix multiplication which would increase memory requirements. 

In Fig.  \ref{patch_attention} we present a simple example of a similarity map,  that consists of the comparison of two binary images of random disks, the query, $q$ - top left panel, and key, $k$ - top middle panel. Each of the images, of spatial size $256 \times 256$, contains three random - in location and size - disks, and the values are binary, where $1$ is assigned on the disk, and $0$ on the background. The query and key are then split in $4 \times 4$ patches of spatial size $64\times 64$, i.e. they are reshaped to $4\times 4 \times 64 \times 64$. The first two $4\times 4$ indices provide information for the coordinates of the patch, while the latter $64\times 64$ are the pixel coordinates of each patch. The spatial ($64\times 64$) parts of these are then  compared to each other using the Tanimoto similarity:
\begin{equation}
\label{Tanimoto_similarity}
\mathcal{T}(q, k) =
\begin{cases}
	 \frac{ \langle \mathbf{q} | \mathbf{ k}  \rangle}{\langle \mathbf  { q}  | \mathbf { q }  \rangle + \langle \mathbf {k} | \mathbf { k } \rangle - \langle \mathbf { q }  | \mathbf { k}  \rangle   } & \mathbf{q} \neq \mathbf{0} \;\text{or}\; \mathbf{k} \neq \mathbf{0}   
\\
\quad \quad 0 & \mathbf{q} = \mathbf{k} = \mathbf{0}  
\end{cases}
\end{equation}
where $\langle \mathbf{q} | \mathbf{ k}  \rangle$ represents the tensor contraction for selected indices of $q$ and $k$. In this particular example these are the spatial indices of each of the $64 \times 64$ image patches. This similarity 
 results in a coordinate map of dimensionality: 
\begin{equation}
\mathcal{T}(q , k) : \underset{\text{query}}{ \underbrace{(4 \times 4)}} \times \underset{\text{key}}{\underbrace{(4 \times 4)}}.
\end{equation}
In this coordinate map, the first two indices correspond to the $4\times 4 $ coordinate patch locations of the query, $q$, and the last two indices to the coordinate spatial patch locations  of the key, $k$. The similarity metric used, is akin to Intersection over Union, thus it measures the overlap of disks in a given patch. Note, we map similarity of empty space (i.e. 0) to 0 with the functional definition of the Tanimoto coefficient. 

For example, from the top right panel of Fig. \ref{patch_attention} we see the similarity of the $q$ patch at location [1,2] (second row, third column): This is a 4x4 image, whose elements are the similarity metrics of the patch [1,2] of $q$ with all the patches of the $k$ image. Therefore the top left pixel is the similarity of $q_{[1,2]} \sim k_{[0,0]}$ and so on. Note that the highest similarity of $q_{[1,2]}$ with $k$-patches,  as highlighted by brighter (white) color in pixel location [1,2], is between patches $q_{[1,2]}$ and $k_{[1,2]}$. In the bottom panel of Fig. \ref{patch_attention} we provide another visual example for the similarity of patch $q_{[2,0]}$ (second column)  and $k$ (third column). Note that the similarity $q_{[2,0]} \sim k$ depicted in the bottom right panel is the third row, first column patch of similarity of top right panel. 
In this spatial map (top right panel of Fig. \ref{patch_attention}) lies all cross spatial correlation information that exist on the query $q$ image. For example, by looking at patches [1,2] and [2,0] of the query, $q$, we see that they consists of two top semi-disks. We anticipate therefore that if a similar structure (i.e. similar sections of disks) appears on the keys, $k$,  image, this will appear as similar heat map on the coordinates of the similarity map, $q \sim k$. Indeed locations (top right panel) [1,2], [1,3], [2,0], [2,2] and [3,0] have brighter colors, suggesting the presence of sections of disks in locations [1,2], [1,3], [2,0] and [2,2] of the query $q$ image. Visual inspection of $q$, verifies this.

Using this simplistic example for comparison of binary images, it becomes also evident that if the discs had color (i.e. channel dimension), then a red disc (represented by (1,0,0) vector, in RGB format) and a blue disk (0,0,1) would have zero similarity, due to their Euclidean dot product being zero. This suggests that a similar patch-splitting in channel dimension is necessary to encapsulate cross channel similarity between images. 

In a more rigorous mathematical treatment (omitting batch dimension for simplicity), assuming dimensionality of $C\times H \times W$ for the query, key and values tensors, i.e. $\mathbf{q}\in \mathfrak{R}^{C\times H\times W}$,  $\mathbf{k}\in \mathfrak{R}^{\text{C}\times  \text{H} \times  \text{W}}$ and $\mathbf{v}\in \mathfrak{R}^{\text{C}\times  \text{H} \times  \text{W}}$ then we can reshape these tensors into $c \times h \times w$ distinct patches: 
\begin{align*}
q_{ C\times H\times W} &\to q_{ c\times h \times w \; \times \; (C/c) \times (H/h) \times (W/w)}\\
k_{C\times H\times W} &\to k_{ c\times h \times w \; \times \;  (C/c) \times (H/h) \times (W/w)} \\
v_{C\times H\times W} &\to v_{ c\times h \times w \; \times \;  (C/c) \times (H/h) \times (W/w)}. 
\end{align*} 
Each of these patches now has channel and spatial dimensions $C/c \times H/h \times W/w$. 
 Writing in index notation the various quantities that participate in the Tanimoto similarity (Eq \ref{Tanimoto_similarity}): 
\begin{align}
	\langle \mathbf{q} | \mathbf{ k}  \rangle &= \sum_{rst} q_{\textcolor{cyan}{chw} rst}   k_{\textcolor{magenta}{klm} rst}  \equiv \langle \mathbf{q} | \mathbf{ k} \rangle {} _{\textcolor{cyan}{chw}\textcolor{magenta}{klm}   }  \\
	\langle \mathbf{q} | \mathbf{ q}  \rangle &= \sum_{rst} q_{\textcolor{cyan}{chw} rst}   q_{\textcolor{cyan}{chw} rst}  \equiv \langle \mathbf{q} | \mathbf{ q} \rangle {} _{\textcolor{cyan}{chw}   } \\	
	\langle \mathbf{k} | \mathbf{ k}  \rangle &= \sum_{rst} k_{\textcolor{magenta}{klm} rst}   k_{\textcolor{magenta}{klm} rst}  \equiv \langle \mathbf{k} | \mathbf{ k} \rangle {} _{\textcolor{magenta}{klm}  } \\
\end{align}
yields that the Tanimoto similarity of $q$ and $k$, has dimensions $\mathcal{T}(q, k)_{ \textcolor{cyan}{chw}\textcolor{magenta}{klm}}$ (see also Listing \ref{qk_similarity}). 

Given that the patches have lower dimensionality than the original channels, $C$, and spatial $H\times W$, dimensions, the similarity matrix $\mathcal{T}(q, k)_{ \textcolor{cyan}{chw}\textcolor{magenta}{klm}}$ does not occupy a significant portion of the memory. To further reduce the memory footprint and  avoid matrix multiplication with the values vector, we summarize the information of all patch comparisons in a single coordinate map. We do so by contracting the first three indices, $\textcolor{cyan}{chw}$ i.e. the query indices, with a \texttt{Linear} weight matrix of dimensionality  $chw \times 1$ (Listing \ref{PTAttentionCODE}, line 33), thus contracting the dimensionality of the coordinate map to $\textcolor{magenta}{klm}$, i.e.:
\begin{equation}
\label{similarity_coordinate_map_weighted}
\tilde{\mathcal{T}}(q , k)_{\textcolor{magenta}{klm}} = \sum_{\textcolor{cyan}{chw}} \mathcal{T}(q , k)_{ \textcolor{cyan}{chw}\textcolor{magenta}{klm}} W_{\textcolor{cyan}{chw}}
\end{equation} 
Finally, the attention layer is produced by element-wise multiplication of this similarity matrix (using the broadcasting technique in python) with the values tensor, $v$, and re-arranging the patches to the original tensor shape, subject to the activation \texttt{d2s} (Section: \ref{d2s_activation}):
\begin{equation}
\label{Attention_map}
\mathcal{A}(q,k,v)_{\textcolor{magenta}{klm}rst} = \texttt{d2s}\left( \tilde{\mathcal{T}}(q , k)_{\textcolor{magenta}{klm}} \odot v_{\textcolor{magenta}{klm}rst} \right)
\end{equation}
The Attention map, $\mathcal{A}(q,k,v)$, upon reshaping, has dimensionality $C\times H\times W$, and encapsulates all spatial and channel correlations that exist between query and key tensors to a patch resolution level of $c\times h\times w$, i.e.  $\mathcal{A}(q,k,v) \in \mathfrak{R}^{C\times H\times W}$  (see also Listing \ref{PTAttentionCODE}).  We name this new attention mechanism Patch Tanimoto Attention (or \texttt{PTA}), and it is a central building block of our feature extraction units.

\subsection{\texttt{d2s} activation function}
\label{d2s_activation}

\begin{figure}[h!]
\centering
  \includegraphics[width=\linewidth]{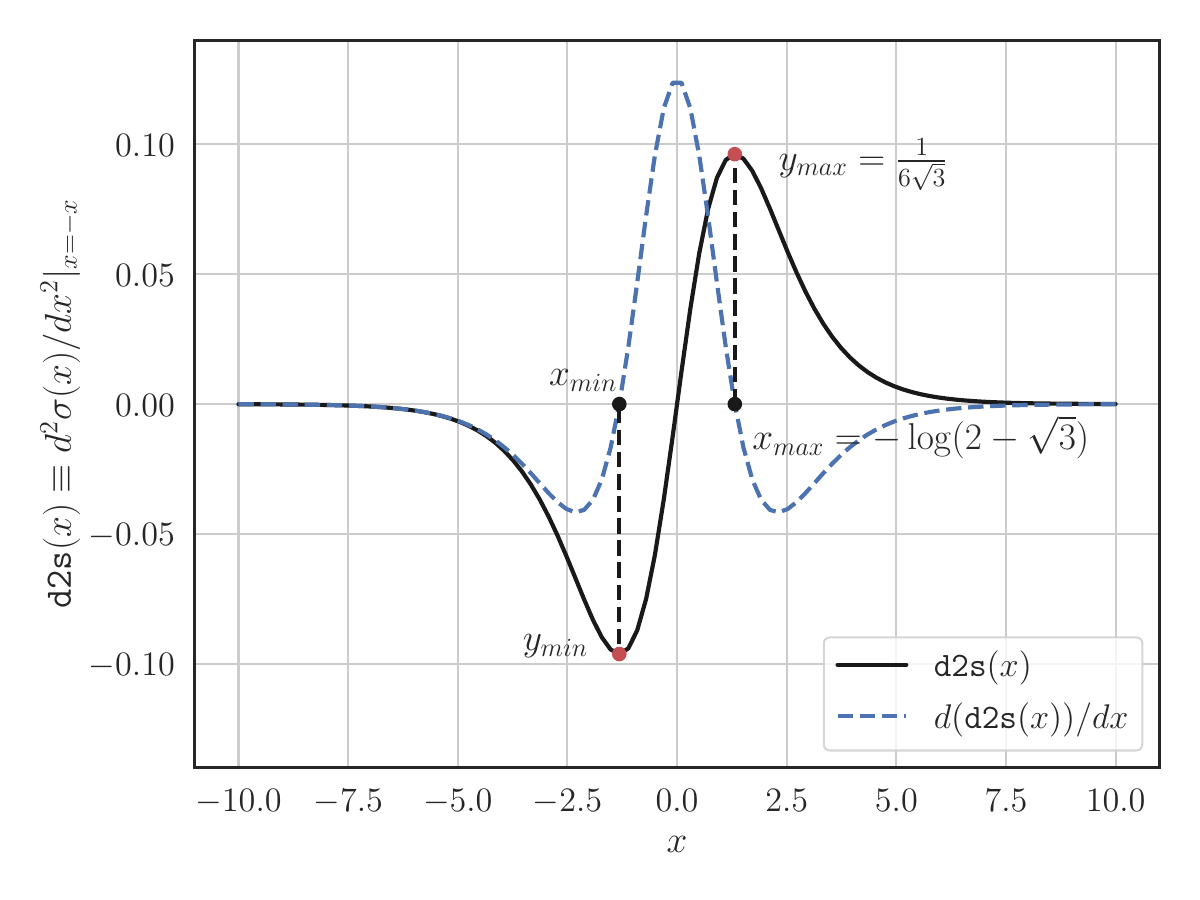}
  \caption{Functional form of the \texttt{d2s} activation function and its derivative.}
  \label{d2sigmoid_act}
\end{figure}

\begin{figure*}[ht!]
    \centering
    \subfigure[PTA-ViT Stage]{%
    \centering
        \includegraphics[clip, trim=0.25cm 0.40cm 0.25cm 0.1cm,width=0.25\textwidth]{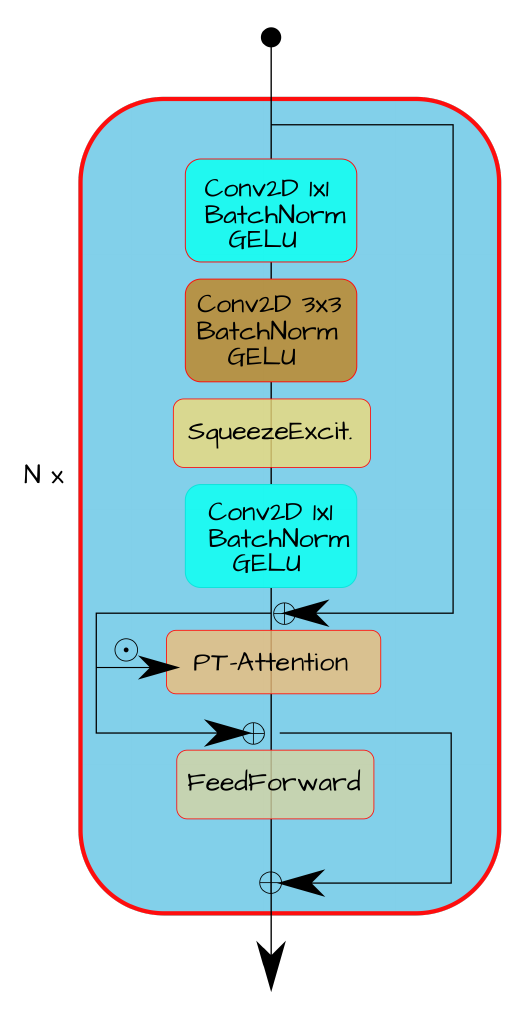}
        \label{ptvitstage:suba}
    }
    \subfigure[U-Net topology]{%
        \includegraphics[clip, trim=0.0cm 0.20cm 0.25cm 0.1cm,width=0.5\textwidth]{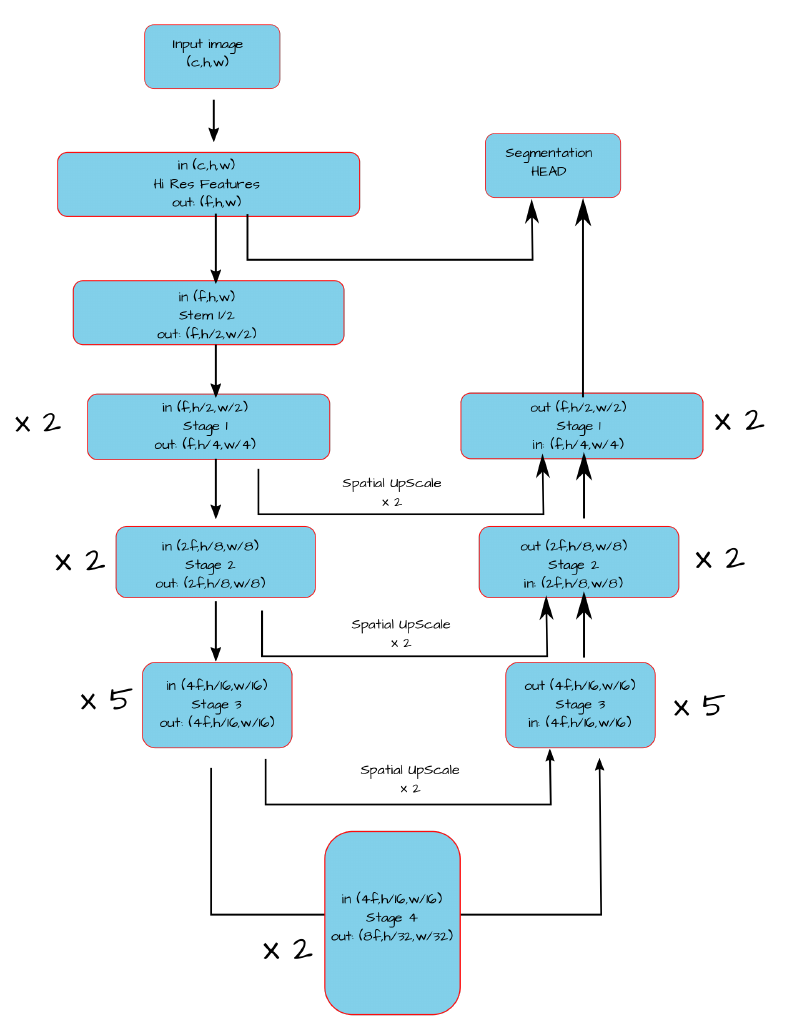}
        \label{ptvitstage:subb}
    }
    \caption{Patch Tanimoto Attention ViT Stage. The \texttt{Stage} comprises a sequence of MBConv blocks, followed by Squeeze Excitation, Patch Tanimoto Attention and a FeedForward network. The architecture is defined by the number of these building blocks that are repeated. The UNet macro-topology is symmetric in its encoder and decoder, which is reflected by the same number of stages used in the corresponding points of the Encoder and Decoder layers. In the Figure we show the Tiny configuration of [2,2,5,2,5,2,2] network.} 
    \label{ptvitstage}
\end{figure*}

We define a new activation function (Fig. \ref{d2sigmoid_act}), the $\texttt{d2s}= \frac{d^2\sigma(-x)}{dx^2} $, given by the functional form:
\begin{equation}
\texttt{d2s}(x) =
2 u^3-3 u^2+u, \; \text{where}\; 
u=\sigma(-x)=\frac{1}{1+\exp(x)}
\end{equation}
This activation function take both positive and negative values within a finite range ($y \in [-0.1,0.1]$), and it avoids the problem of vanishing gradients. This is because it does not attain an asymptotic constant value for some x larger or smaller than some threshold (as \texttt{sigmoid} and \texttt{tanh} activation functions do). In fact, the maximum and minimum values, $y_{\max/\min}\pm 3^{1/2}6^{-1}$, occur at finite locations $x_{\min/\max}\pm \log(2-\sqrt{3})$. Therefore, for maximum/minimum activation values, the numerical values are encouraged to be around this range. In Fig. \ref{d2sigmoid_act} we plot the \texttt{d2s} activation function (solid line) along with its derivative (dashed line). It is seen that around local minima/maxima the gradients do not stagnate to asymptotic values. Additionally, the magnitude of the derivative is comparable to the value of the activation function.  The \texttt{d2s} activation function is exclusively used within the \texttt{PTA} attention module.

\subsection{Feature extraction units}

The foundation of our feature extraction units builds on MaXViT \citep{tu2022maxvit}. In our approach, we replace the sequence of block and grid attention units with a single patch attention unit. Our implementation follows the ViT architecture as detailed in \href{https://github.com/lucidrains/vit-pytorch/blob/main/vit_pytorch/max_vit.py}{lucidrains/vit\_pytorch.git}. We refer to these modified feature extraction units as \texttt{PTA-ViT} (Patch Tanimoto Attention ViT) and to  the sequential stack of such units as a \texttt{Stage} (Fig. \ref{ptvitstage:suba}).

\subsubsection{Prelude: fuzzy set intersection and union}

In fuzzy set theory, the concepts of set intersection and union are generalized to accommodate membership degrees that range between 0 and 1, as opposed to just binary membership. While the  $\min / \max$ functions are commonly used, alternative t-norms and t-conorms can also be employed for intersection and union, respectively.

Let $\tnorm$ be a t-norm that represents intersection in fuzzy set theory. Then the union $\conorm$ is defined as 
\begin{equation}
\conorm(a,b) = 1- \tnorm(1-a,1-b)
\end{equation}
We adopt the  \cite{hamacher1978uber} product as t-norm to better mix the gradients of the predictions (in contrast if we were to use the $\min$ operation, only one of the two layers, $a$ or $b$ would contribute to the gradient calculation): 
\begin{equation}
\tnorm \equiv H_{\gamma}(a,b) = \frac{a b}{\gamma + (1-\gamma)(a+b-ab)}.
\end{equation}
Here we set $\gamma$ to \texttt{1.e-5} to avoid division by zero. In the following for simplicity we will use the standard set operations symbols,  $\cap \equiv \tnorm$ and $\cup \equiv \conorm$ to avoid confusion to the reader, however it should be understood that wherever these operations are applied to the predictions of the network, these correspond to the fuzzy operations defined above.  

The set difference operation is defined as the elements in a that are not in b, e.g. if $a = \{1,2,3\}$ and $b=\{2,3,4,5\}$ then $a\setminus b=\{1\}$. This can be expressed with the aid of the intersection and set complement operations: 
\begin{equation}
a\setminus b = a \cap (\overline{a\cap b})
\end{equation} 
where $\overline{a\cap b}$ is the complement of the set intersection $a\cap b$. 

\subsection{UNet macro-topology}

We construct a basic UNet \citep{DBLP:journals/corr/RonnebergerFB15} macro-topology, in which we replace the standard feature extraction units with PTA-ViT stages. The details of this implementation can be seen in  Fig \ref{ptvitstage:subb}. We note that in this architecture, the encoder and decoder are symmetric, as this is depicted by the same number of stages used in the encoder and the decoder. This will be the ``base'' architecture against which we will compare the SSG2 modelling framework.

\subsection{SSG2 Model architecture}

The model architecture is broken into two components. A base model that takes as input two images and produces a single segmentation mask that corresponds to the set intersection of the ground truths of these input images. And a sequence module that is able to consume and process a set  of features. 
This sequence of features is generated by the base model, where the first input - the target image - is kept fixed, while the second input image - the support image - is randomly selected.

\begin{figure}
\centering
\includegraphics[clip, trim=2.25cm 6.40cm 0.25cm 5.1cm,width=\columnwidth]{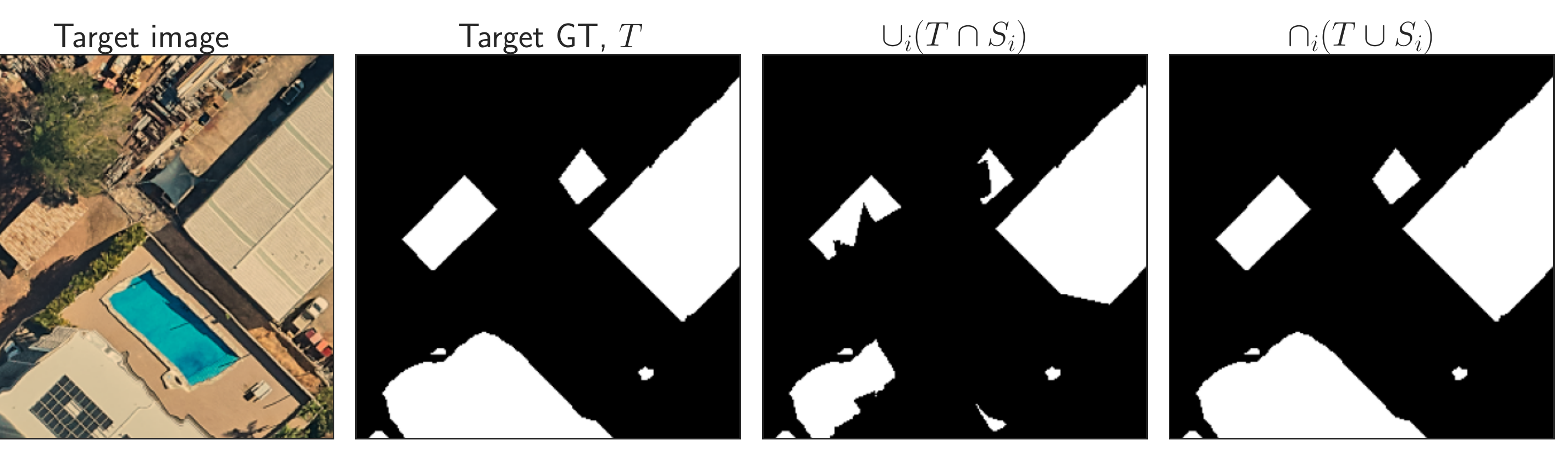}
\caption{From left to right: target image, ground truth of target image, fuzzy set union of the intersection ground truths and intersection of the unions. This is for a sequence length of 16 support images. } 
\label{example_target_image_vs_gt_vs_fz_set_union}
\end{figure}

The Sequence model then consumes the (set of) extracted features from the decoder, as well as the fuzzy set union of the base model predictions of intersections, and the fuzzy set intersection of the union predictions (Fig. \ref{system_sequence_model}). From these it produces a (multitasking) segmentation mask that corresponds to the target image ground truth. That is, the sequence model processes a set of features, attempting to discriminate between erroneous predictions and true features. The key idea here is that true features will have different geometric shape and frequency of appearance in comparison with features that correspond to noise. In other words, adding the extra sequence dimension helps distinguish between noise and true features. 
The fuzzy union of intersections, $\bigcup_{i\in I}(T\cap S_i)$, and intersection of unions, $\bigcap_{i\in I}(T\cup S_i)$, is a very strong prior (Fig. \ref{example_target_image_vs_gt_vs_fz_set_union}) for the full segmentation mask. It will also allow to discriminate between genuine features and noise.

In the following we proceed by first detailing the inner workings of the base model, then the Sequence model and finally the overall architecture.

\subsubsection{Base model feature extraction unit}

\begin{figure*}[!ht]
\centering
\includegraphics[clip, trim=1.25cm 0.40cm 1.5cm 2.1cm,width=\textwidth]{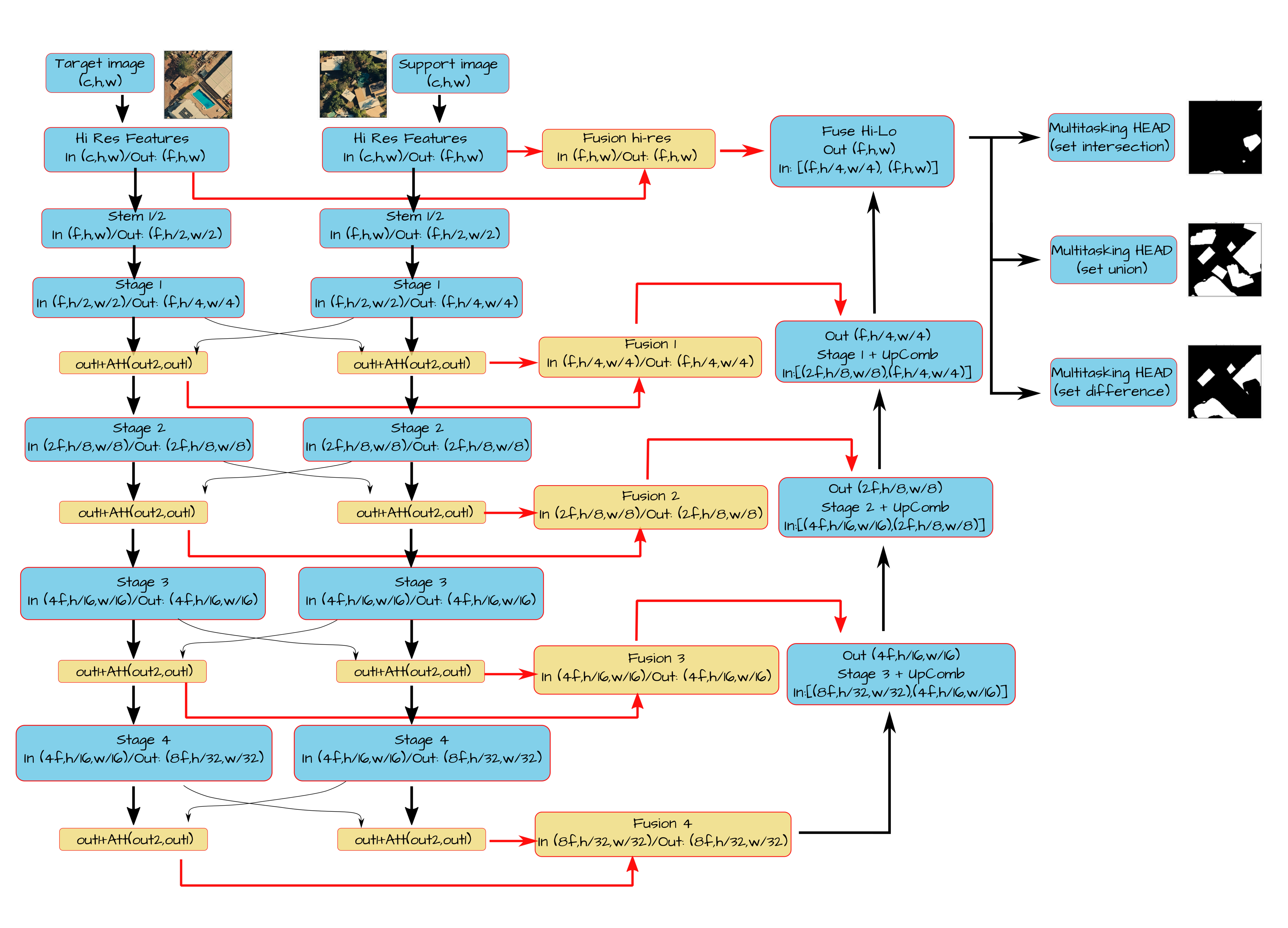}
\caption{Base architecture of set intersection, union, and difference predictions. } 
\label{mantis_v2_base_arch}
\end{figure*}

The base model, that consumes as input two images and predicts a segmentation mask that corresponds to the set intersection of the ground truths of each individual input can be seen in Fig. \ref{mantis_v2_base_arch}. The basic macro topology, inspired from Siamese networks \citep{Koch2015SiameseNN,c441a21f627f4b0a896fa62cb143176f}
 is borrowed from \citep{rs13183707}, with the following differences: (a) for \texttt{Stages} we use the \texttt{PTA-ViT} feature extraction units; (b) for the Fusion parts of the architecture, we use the same Fusion technique as in \citep{rs13183707} but now the attention weights are common for both cross (or relative) attention layers, and the attention mechanism is the one developed here. Finally, (c) the relative (cross) attention is passed into each branch of the encoder as it can be seen in Fig \ref{mantis_v2_base_arch}. 

One important feature in this architecture for memory efficiency, is the recipe from ViT and variants of the aggressive spatial reduction in the first  layer of the encoder, i.e. even before they enter the encoder/decoder structure of the model. In our case this happens in the \texttt{Stem} layer and \texttt{Stage 1} of the encoder. This allows for significantly higher number of features on the layers of the architecture in comparison with the traditional 16 or 32 of the U-Net   \citep{DBLP:journals/corr/RonnebergerFB15} (and variants) architectures. This, however, can also lead to reduced efficiency due to different resolution of the extracted features from the last Stage (1/4 of the input resolution). We tackle this problem, by combining  an initial full resolution layer with the extracted features, before including this into the multitasking head. 

The hyper-parameter selection we follow is the one from Tiny ViT due to the memory footprint of the whole structure, we have 4 stages ([2,2,5,2]) for the encoder, and 3 for the decoder ([5,2,2]), starting from 96 initial features, which are doubled in every subsequent stage of the encoder architecture, and then halved for each subsequent stage of the decoder.

\subsubsection{Sequence Model}

For Sequence model we create a 3D realization of the \texttt{PTA-ViT} stage, termed \texttt{PTA-ViT3D}, that replaces the 2D convolutions with 3D ones and is also equiped with a patch attention module tailored for 3D sequence data. In particular, for an input tensor of shape (we omitt the batch dimension) $q_{C \times S \times H \times W}$, we split it in $h,w$ spatial patches, but we keep the whole partitioning of the sequence dimension. That is, we demand this attention to compare all sequence elements with each other for each image patch, there is no partitioning in sequence dimension.  The $q$, $k$ and $v$ dimensionality in this case becomes: 
\begin{align*}
q_{ C\times S\times H\times W} &\to q_{  S\, \times \, h \, \times \, w \; \times \; C \times \, (H/h)\, \times \, (W/w) } \\
k_{ C\times S\times H\times W} &\to k_{  S\, \times \, h \, \times \, w \; \times \; C \times \, (H/h)\, \times \, (W/w) }\\
v_{ C\times S\times H\times W} &\to v_{  S\, \times \, h \, \times \, w \; \times \; C \times \, (H/h)\, \times \, (W/w) },
\end{align*} 
where $S$ is the length of the sequence, and $h$, $w$ the horizontal and vertical number of image patches.
Therefore, the similarity between $q$ and $k$ tensors, reflects all sequence correlations. The summation is with respect to the full channel space, $C$, and the spatial patches, $H/h$ and $W/w$. We do not split on channel spaces, because the extracted features are already refined, therefore, there will be channel alignment before the final segmentation head.  The dimensionality of the similarity between $q$ and $k$ is now
$\langle \mathbf{q}, \mathbf{k} \rangle \to S \times h  \times  w \times S  \times  h  \times  w.$ Then, in a process similar to Equations \eqref{similarity_coordinate_map_weighted} and \eqref{Attention_map} we construct the attention that is applied to the 3D features. 

The role of the \texttt{PTA-ViT3D} stage is to consume a 3D tensor of features, $F_{cshw}$, and correlate spatially, $h,w$, as well as across the sequence dimensions the features. The output of this 3D module must be a set of 3D features, which are similar in all sequence slots. That is, the signature of set operations is erased, and is replaced with the features corresponding to the target image. We designate these features with the symbol $\mathcal{F}_{cshw}$. The abelian nature with respect to the sequence index is established via random permutation in the loss function. 

We detail the usage of the sequence modelling in Listing \ref{headcmtsk3d} (see also Fig. \ref{system_sequence_model}). There, \texttt{head\_cmtsk} corresponds to the standard conditioned multitasking head for the prediction of segmentation, boundaries and distance  as it was introduced in \citet{rs13183707}.   However, one important difference exists in the number of classes when training the SSG2 model: we introduce an auxiliary dimension termed the ``null class''. This class serves a specific role during the set operations -- intersection, union, and difference -- performed at each step of the sequence model. Unlike the ``background'' class, which represents actual background regions in the image, the ``null class'' is used to denote the absence of any class label between two images when performing set operations. This distinction proves critical when using 1-hot encoding, where the ``null'' class effectively accommodates regions that result from set operations as having no corresponding class label. It is worth noting that this is an optional modification; however, we employ it in our work to minimize potential bias that may arise from conflating the ``null'' class with the ``background'' class, especially given the variable frequency presence of the latter. During inference, the null class dimension is removed, reverting to the standard 1-hot encoding scheme for class labels.

\subsection{Loss function}

\subsubsection{Loss functional form}
For the loss function we use for all layers the multitasking approach we developed in \citep{DIAKOGIANNIS202094}, i.e. the Tanimoto with complement: 
\begin{equation}
\mathcal{L}_{\mathcal{T}}(\mathbf{p},\mathbf{l})= 1 -\frac{1}{2} \biggl(\mathcal{T}(\mathbf{p},\mathbf{l}) + \mathcal{T}(1-\mathbf{p},1-\mathbf{l})\biggr)
\end{equation}
for the segmentation, distance transform and boundaries of each object. In addition to these we augment the learning with some theoretical constraints (implemented as loss function terms) from set theory. Namely, for each pair of target, $T$, and support, $S_j$, images and their corresponding ground truths, $l_T, l_{S_j}$, in the $j$ sequence we demand: 
\begin{align}
\hat{P}_{T\cap S_j} \cup \hat{P}_{T \setminus S_j} & \to l_T \\
\hat{P}_{T\cap S_j} \cap \hat{P}_{T \setminus S_j} & \to l_\varnothing  %
\end{align}
 where $\hat{P}$ correspond to the predictions of the network for each set and $l_\varnothing$ is a vector of zeros corresponding to the null set. We denote $\sigma_s\mathcal{F}_{cshw}$   a random permutation of the features along the sequence dimension of the output of the \texttt{PTA-ViT3D} stage. 
   The total loss is therefore:
\begin{align}
\notag 
\lossT(p,l) &= \lossT(\hat{P}_T,l_T) +   \sum_{j\in I} \biggl\{
\lossT(\hat{P}_{T\cap S_j},l_{T\cap S_j})
\\ \notag  &+
\lossT(\hat{P}_{T\cup S_j},l_{T\cup S_j})
+
\lossT(\hat{P}_{T\setminus S_j},l_{T\setminus S_j})
\\ \notag &+ 
\lossT(\hat{P}_{T\cap S_j} \cup \hat{P}_{T \setminus S_j} , l_T)
\\&+ 
\lossT(\hat{P}_{T\cap S_j} \cap \hat{P}_{T \setminus S_j} , l_\varnothing) \biggr\}
\\ &+
\lossT(\sigma_s\left(\mathcal{F}_{bcshw}\right),\mathcal{F}_{bcshw})
\end{align} 
where the summation $j$ extends to all the support images and $\hat{P}_T$ corresponds to the predictions of the Target image from the sequence modelling.

\subsection{Data pre-processing and Augmentation}

For the UrbanMonitor and ISPRS datasets, in the pre-processing phase, the dataset undergoes standardization to attain zero mean and unit standard deviation. To accommodate GPU memory limitations, we extract overlapping training chips with a window size of $256 \times 256$   pixels and at a stride of 128 pixels. This approach ensures that the training chips are of a manageable size to fit into the available GPU memory. 

 To bolster the model's generalization performance, we employ a composite suite of data augmentation techniques. These are orchestrated through a probabilistic framework that selects one of several geometric transformations to apply to each image. Specifically, the transformations are executed with equal probability and encompass horizontal and vertical flips, elastic transformations that offer perspective projection, grid distortions with a distortion limit of 0.4, and shift-scale-rotate operations with a shift limit of 0.25, a scale limit ranging from 0.75 to 1.25, and a rotation limit of 180 degrees.

 For the ISIC2018 dataset, images are resized to 256 $\times$ 256 pixels, preserving the original aspect ratio through zero-padding in the minor dimension as required. During training, we again follow the same data augmentation methodology as in the remote sensing data as the aim here is to see if the algorithm developed can work in a cross discipline manner without any modifications.

 This multifaceted augmentation strategy serves to enrich the training dataset, thereby enhancing the robustness and adaptability of our deep learning architecture. The transformations were realized using the library Albumentations \citep{info11020125}.

\subsection{Evaluation Metrics}

All of the evaluation metrics were based on calculation first of the confusion matrix with the use of the package \texttt{PyCM} \citep{Haghighi2018-2}. For the evaluation of the performance,   we use the following metrics:

\subsubsection{Matthews Correlation Coefficient}
 The Matthews Correlation Coefficient  \citep{MATTHEWS1975442} in its multiclass version \citep{GORODKIN2004367}, defined by a $K\times K$ confusion matrix, $C_{ij}$, where $K$ is the number of classes, is given by:
\begin{equation}
\label{mcc_multiclass}
MCC = \frac{c s - \sum_{i=1}^K p_i t_i }{\sqrt{(s^2-\sum_{i=1}^K p_i^2) (s^2-\sum_{i=1}^K t_i^2)}}
\end{equation}
where 
\begin{align*}
t_i &= \sum_{j=1}^K C_{ji} \quad \text{ represents the actual occurrence count of class $k$,}\\
p_i &= \sum_{j=1}^K C_{ij}\quad \text{indicates how many times class $k$ was predicted,}\\
c   &= \sum_{i=1}^K C_{ii} \quad \text{is the total number of correct predictions,} \\
s   &= \sum_{i=1}^K \sum_{j=1}^K C_{ij} \quad \text{is the overall sample count.}
\end{align*}
MCC ranges from [-1,1] in the binary case, where a value of 1 suggests maximum performance. For the multiclass case the lower value $\in$ [-1,0].

\subsubsection{Cohen's kappa}

Cohen's Kappa $\kappa$ is a statistical measure used to assess the reliability of categorical classifications made by multiple raters. Unlike simple accuracy, Cohen's Kappa accounts for the possibility of agreement occurring by chance, thus providing a more robust evaluation of classification performance. It is calculated as:
\begin{equation}
\kappa = \frac{P_o-P_e}{1-P_e}
\end{equation}
where $P_o$ is the observed agreement and $P_e$ the expected agreement. In terms of the confusion matrix, $C_{ij}$, for $K$ classes it can be defined as: 
\begin{equation}
\kappa = \frac{c s - (\sum_{i=1}^K  p_i t_i)}{s^2-(\sum_{i=1}^K  p_i t_i)}
\end{equation}
where the various quantities $c,s,t_i$ and $p_i$ are the same as used in the MCC definition. 

\subsubsection{Bookmakers Informedness and Markedness}

 In the context of binary classification, Bookmakers Informedness (BM) and Markedness (MK) \citep{6b87510ce7324df69116f8395644ed77} are defined via Sensitivity (Recall or True Positive Rate - TPR) and Specificity (True Negative Rate - TNR) as:
\begin{equation}
\mathcal{BM} = \text{TPR} + \text{TNR} - 1
\end{equation}     
Informedness shows how informed a model is, compared to random guessing.  It  provides a balanced measure that considers both the true positive rate (Sensitivity) and the true negative rate (Specificity). This metric effectively evaluates how ``informed'' a decision-maker, metaphorically referred to as a ``bookmaker'', would be if they based their decisions on the model's predictions. It offers a more balanced view and is generally less sensitive to class imbalance, making it a robust choice for generalized performance assessment.

Markedness is a metric on how reliable the predictions are, calculated as: 
\begin{equation} 
\mathcal{MK} = \text{PPV} + \text{NPV} - 1 
\end{equation}
(where PPV is Positive Predictive Value and NPV is Negative Predictive Value). Unlike Precision, which focuses solely on the quality of positive predictions, Markedness accounts for both the positive and negative predicted classes, thereby providing insights into the model's overall reliability. It takes into account both false positives and false negatives, making it a valuable metric when assessing the trustworthiness of a model.

Both metrics take values in the range [-1,1], similar to the MCC. 

\subsubsection{Mean Intersection over Union}

For two one-hot encoded binary predictions of shape $N\times H  \times W$, where $N$ is the number of classes, $H$ and $W$ the height and width respectively, $P$ and $L$ we define the mean Intersection over Union, via the (fuzzy) set operations of intersection and union as: 
\begin{equation}
\text{mIoU} = \frac{1}{N}\sum_{i=1}^N \frac{\sum_{j,k}\text{min}(P_{ijk},L_{ijk})}{\sum_{j,k}\text{max}(P_{ijk},L_{ijk})}
\end{equation}

For binary one dimensional masks, we use the following definition:
\begin{equation}
\text{IoU} = \frac{TP}{TP+FP+FN}
\end{equation}

\subsubsection{Dice Coefficient}
For the case of the ISIC2018 dataset, we also evaluate the  \cite{doi:10.2307/1932409} coefficient:
\begin{equation}
\label{dice_formula}
\text{Dice} = \frac{ 2  TP}{ 2  TP + FP + FN}
\end{equation}

\begin{figure}
\centering
\includegraphics[clip, trim=0.25cm 0.40cm 0.25cm 0.1cm,width=\columnwidth]{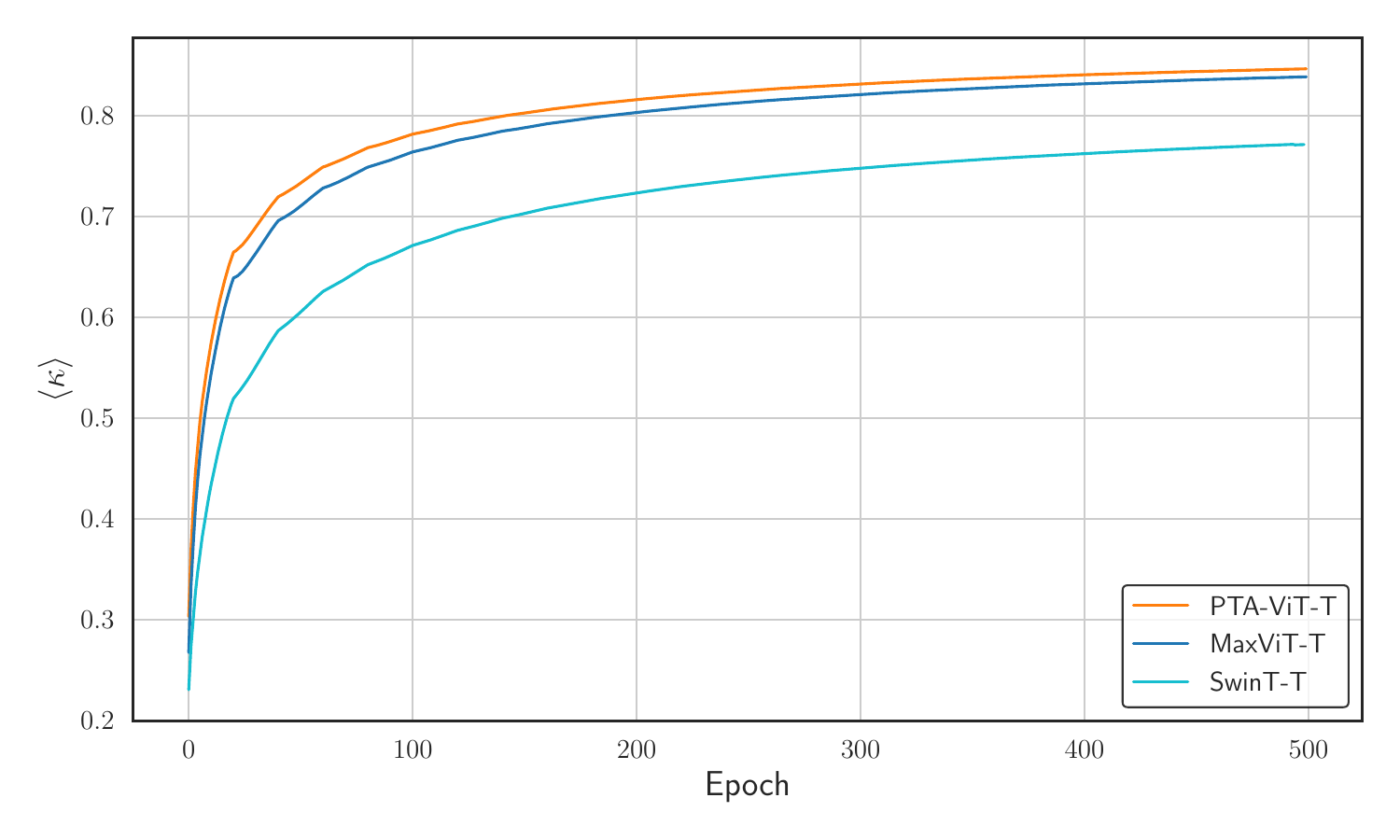}
\caption{Comparison of MaxViT-T architecture, with this work (PT-ViT-T) on CIFAR10. The two models differ only in the Attention mechanism, where in our work we use the Patched Tanimoto Attention.} 
\label{cifar10_comparison_maxvit_t}
\end{figure}

\subsection{ Experimental Design} 
Our experimental design aims to systematically address a range of research questions concerning the efficacy and versatility of our framework. While our primary area of expertise is remote sensing, we also aim to explore the cross-disciplinary potential of our algorithm. This is in light of the fact that our previous work, ResUNet-a \citep{DIAKOGIANNIS202094}, has found applications in the medical imaging field. With this broader context set, we go on to address the following research questions:
\begin{enumerate}
\item How does the newly introduced attention mechanism compare with the original MaxViT architecture? To assess this, we conduct experiments comparing \texttt{PTA}-\texttt{ViT}-T with MaxViT-T and Swin-T on the CIFAR-10 dataset. All models are tested in their tiny configurations for reasons of computational efficiency. The detailed validation is presented in Section \ref{valid_ptattention}.
\item Does the novel activation function mitigate the issues of vanishing and exploding gradients? This aspect is explored in Section \ref{valid_d2s}, where we compare the training performance of models utilizing the new \texttt{d2s} activation function with those employing a standard sigmoid activation.
\item How does our algorithm fare on semantic segmentation tasks when compared to state-of-the-art architectures? We perform evaluations on the ISPRS dataset, a well-curated benchmark, to offer comprehensive comparisons. The results are discussed in Section \ref{valid_sota_isprs}.
\item What is the efficacy of the proposed SSG2 approach in comparison to UNet-like architectures, particularly when applied to datasets with varying densities? To this end, we focus on understanding the influence of the number of support elements during training. We conduct experiments using the UrbanMonitor (Darwin) dataset and employ three distinct configurations. The first employs a single training tile and five test tiles with a training patch size of 256\( \times \)256 (hereafter F256). The second configuration also uses F256 but includes four training tiles and two test tiles. Finally, the third configuration utilizes a smaller training patch size of 128\( \times \)128 (hereafter F128) with one training tile and five test tiles, allowing for an increased number of support elements during training. These configurations are further discussed in Section \ref{valid_richpoor_sequence}.
\item How does the number of sequence elements influence the model's performance during inference (Section \ref{section_effect_of_support_elements})?
\item What insights can be gleaned from the model's set operations, specifically intersection, union, and difference? The features of key components are visualized and discussed in Section \ref{valid_set_operations}.
\item To build upon the cross-disciplinary utility mentioned earlier, we test our algorithm on the ISIC2018 dataset \citep{DBLP:journals/corr/abs-1902-03368,tschandl2018ham10000}, focusing on skin lesion segmentation, particularly melanoma. 
\end{enumerate}

\section{Results}

\subsection{Validation of the PT-Attention}
\label{valid_ptattention}

We took the original MaxViT-T architecture\footnote{We used the implementation from \href{https://github.com/lucidrains/vit-pytorch/blob/main/vit_pytorch/max_vit.py}{\texttt{lucidrains}} accessed on Dec 2022.} and replaced the multi-axis attention blocks, with our own Patch Tanimoto Attention. We run for 500 epochs, with the same hyperparameters and learning rate scheduler (Cosine Annealing with linear warmup strategy). For completion we also compare with a version of the SwinTransformerV2\footnote{Official implementation from  Microsoft \href{https://github.com/microsoft/Swin-Transformer/tree/2cb103f2de145ff43bb9f6fc2ae8800c24ad04c6}{repository}.}. The result of the experiment can be seen in Fig.  \ref{cifar10_comparison_maxvit_t} and Table \ref{tab:comparison_cifar10}, where the new attention provides better performance to the otherwise similar ViT blocks with less parameters. The PTA-ViT-T has $\sim$33\% smaller memory footprint for this model (estimated total size for ($32\times 32 \times 3$ input image), less parameters than both MaxVit-T and SwinT yet better performance. For the SwinTransformerV2 we used window size = 2 and patch size = 2, given the input image is 32 (almost 8 times smaller than the default 224 \texttt{SwinT} is built for that uses window size=7).

\begin{table}
    \centering
    \fontsize{7}{11}\selectfont
    \begin{tabular}{lccc}
        \toprule
        & \texttt{PTA-ViT-T} & \texttt{MaxViT-T} & \texttt{SwinT} \\
        \midrule
        \textbf{Total Parameters (M)} & 21 & 35 & 27.5 \\
        \textbf{Total Mult-Adds (M)} & 360.93 & 410.78 & 52.64 \\
        \textbf{Input Size (MB)} & 0.02 & 0.02 & 0.02 \\
        \textbf{Fwd/Bwd Pass Size (MB)} & 50.69 & 61.39 & 16.92 \\
        \textbf{Params Size (MB)} & 84.76 & 141.19 & 84.39 \\
        \textbf{Estimated Total Size (MB)} & 135.48 & 202.61 & 101.34 \\
        \hline
        \textbf{MCC @ Epoch 500} & \textbf{84.63 $\pm$ 0.07}\% & 83.84 $\pm$ 0.05 \% & 77.12 $\pm$ 0.28 \% \\
        \bottomrule
    \end{tabular}
    \caption{Comparison of \texttt{PTA-ViT-T}, \texttt{MaxViT-T} and \texttt{SwinTV2-T} on CIFAR-10.}
    \label{tab:comparison_cifar10}
\end{table}

\subsection{Validation of the D2S activation function, for deep networks}
\label{valid_d2s}

\begin{figure}
\centering
  \includegraphics[width=\linewidth]{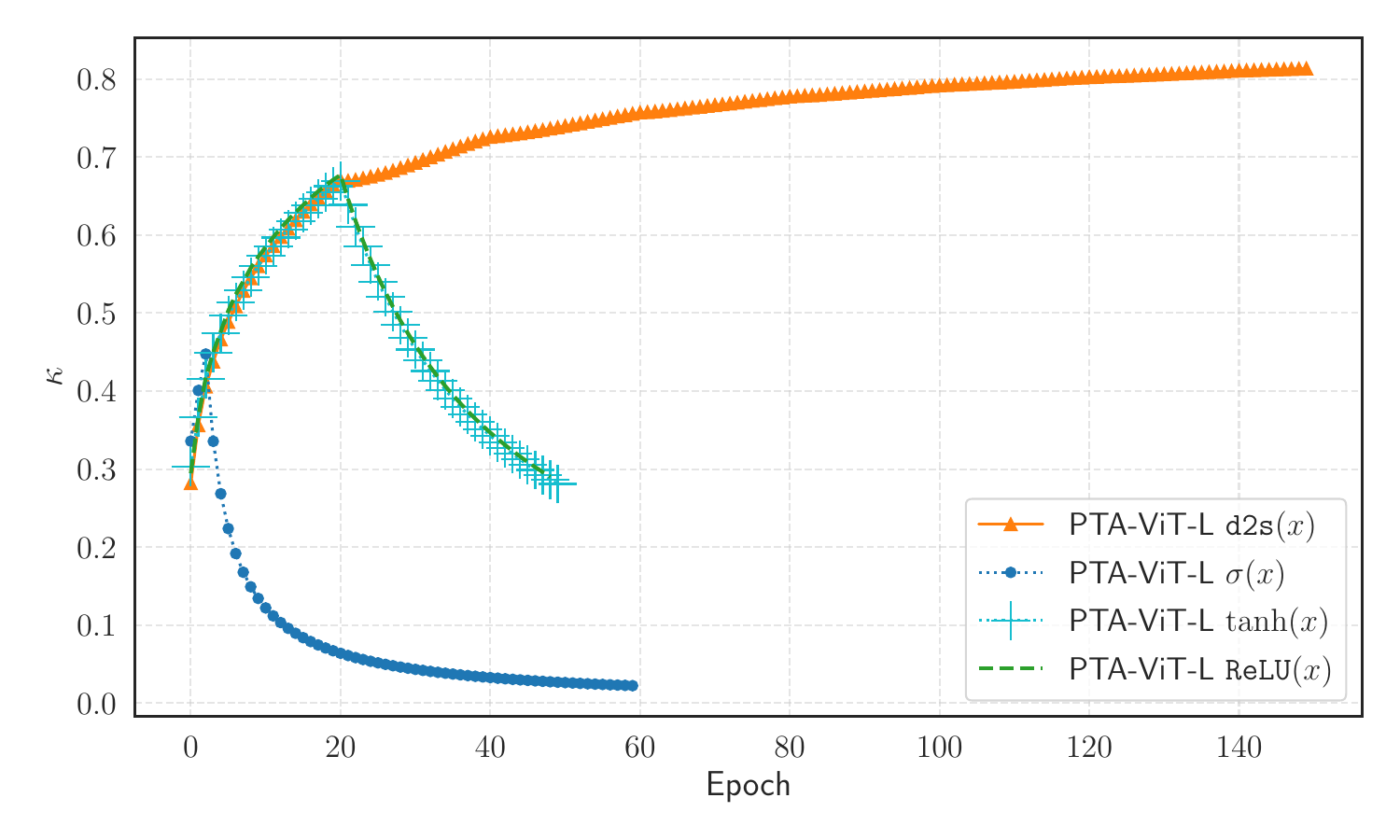}
  \caption{Comparison of the training evolution for \texttt{PTA-ViT} Large with $\sigma(x)$, $\tanh(x)$ and \texttt{ReLU} vs \texttt{d2s} activation function. The models with $\sigma(x)$, $\tanh(x)$ and \texttt{ReLU} activations fail to converge.}
  \label{d2sigmoid_act_vs_sigmoid_PTA_ViT_L}
\end{figure}

\begin{table*}[h]
\centering
\caption{Comparison of results on the ISPRS Potsdam Dataset, for the test tiles, excluding the boundaries. The mIoU as well as $\langle F1 \rangle$ are calculated by excluding the background class, while Overall Accuracy (OA) includes it.}
\label{isprs_results_summary}
\fontsize{8pt}{8pt}\selectfont
\begin{tabularx}{\linewidth}{l l c c c c c c c c}
\toprule
Method & Backbone & Imp.surf. & Building & Low. veg. & Tree & Car & MeanF1 & OA & mIoU \\
\midrule
DST\_5 \citep{Sherrah2016} & FCN & 92.5 & 96.4 & 86.7 & 88.0 & 94.7 & 91.7 & 90.3 & – \\
V-FuseNet \citep{audebert2018beyond} & FuseNet & 92.7 & 96.3 & 87.3 & 88.5 & 95.4 & 92.0 & 90.6 & – \\
UFMG\_4 \citep{8727958}  & – & 90.8 & 95.6 & 84.4 & 84.3 & 92.4 & 89.5 & 87.9 & – \\
S-RA-FCN (Mou et al., 2020) & VGG16 & 91.3 & 94.7 & 86.8 & 83.5 & 94.5 & 90.2 & 88.6 & 82.4 \\
HUSTW4 (Sun et al., 2019) & ResegNets & 93.6 & 97.6 & 88.5 & 88.8 & 94.6 & 92.6 & 91.6 & – \\
TreeUNet (Yue et al., 2019) & – & 93.1 & 97.3 & 86.8 & 87.1 & 95.8 & 92.0 & 90.7 & – \\
ResUNet-a \citep{DIAKOGIANNIS202094} & – & 93.5 & 97.2 & 88.2 & 89.2 & 96.4 & 92.9 & 91.5 & – \\
DDCM-Net (Liu et al., 2020) & ResNet50 & 92.9 & 96.9 & 87.7 & 89.4 & 94.9 & 92.3 & 90.8 & – \\
LANet (Ding et al., 2021) & ResNet50 & 93.1 & 97.2 & 87.3 & 88.0 & 94.2 & 92.0 & 90.8 & – \\
AFNet (Yang et al., 2021b) & ResNet50 + 18 & \textbf{94.1} & 97.6 & 88.7 & 89.7 & \textbf{97.1} & 93.4 & 92.1 & – \\
HMANet (Niu et al., 2021) & ResNet101 & 93.9 & 97.6 & 88.7 & 89.1 & 96.8 & 93.2 & 92.2 & 87.3 \\
STransFuse (Gao et al., 2021) & – & 89.8 & 93.9 & 82.9 & 83.6 & 88.5 & 82.1 & 86.7 & 71.5 \\
SwinB-CNN + BD (Zhang et al., 2022) & Swin-Base & 92.2 & 95.3 & 83.6 & 89.2 & 86.9 & 89.4 & 90.4 & – \\
SwinTF-FPN (Panboonyuen et al., 2021) & Swin-Small & 93.3 & 96.8 & 87.8 & 88.8 & 95.0 & 92.3 & 91.1 & 85.9 \\
ResT (Zhang and Yang, 2021) & ResT-Base & 92.7 & 96.1 & 87.5 & 88.6 & 94.8 & 91.9 & 90.6 & 85.2 \\
FT-UNetFormer \citep{Wang_2022} & Swin-Base & 93.9 & 97.2 & \textbf{88.8} & \textbf{89.8} & 96.6 & 93.3 & 92.0 & 87.5 \\[7pt]
\hline\hline \\[1pt]
\texttt{PTA-ViT-T} (UNet) @ e100 & - & 93.4 & 97.4&  87.6 &  88.6 & 96.3 & 92.7 & 91.1 & 86.6 \\
\texttt{PTA-ViT-T} (UNet) @ e521 & - & 93.4& 97.3& 87.8 & 88.7  & 96.5 &92.7 & 91.1 & 86.7\\
\texttt{PTA-ViT-T} SSG2 (NSupport=5)  @ e117 & - & 93.6 & 97.6 & 88.3 & 88.7 & 96.5 & 92.9 & 91.5 & 87.1\\[5pt]
\texttt{PTA-ViT-T} SSG2 (NSupport=5)  AVG & - & 93.7 & \textbf{97.7} & 88.4 & 88.9 & 96.8 & 93.1 & 91.6 & 87.3\\
\bottomrule
\end{tabularx}
\end{table*}

In experiments with the PTA-ViT-L Large model featuring [2,6,14,2] stages, we observe divergent behavior among different activation functions. Specifically, networks utilizing sigmoid, tanh, and \texttt{ReLU} activations in the PT-Attention mechanism encounter issues with non-convergence, as evidenced by the emergence of \texttt{NAN} values (Fig.  \ref{d2sigmoid_act_vs_sigmoid_PTA_ViT_L}). The \texttt{d2s} activation, however, sidesteps this issue, indicating improved stability in training.

Addressing the challenge of exploding gradients commonly involves the use of gradient clipping, which introduces its own complexities. These include the additional task of hyperparameter tuning for the clipping threshold and potential distortion of the learning signal (due to loss of information from gradient truncation). The \texttt{d2s} activation alleviates these issues by eliminating the need for gradient clipping altogether. This results in a more straightforward hyperparameter setup and maintains the integrity of the gradient information, thereby offering a more streamlined and reliable approach to training deep neural networks.

While the \texttt{d2s} activation demonstrates notable stability in training, it is crucial to consider that the performance of any activation function is tightly interwoven with the choice of weight initialization schemes \citep{pmlr-v9-glorot10a,pmlr-v28-sutskever13,7410480}. The synergy between activation functions and initialization schemes is a complex landscape that can significantly influence not only convergence but also the generalization of the network. For a more rigorous evaluation of \texttt{d2s}, extensive experiments with various initialization strategies are essential. Only then can we fully understand the activation's potential and limitations in diverse architectures and tasks. Future work should aim to examine this interplay between activation functions and initialization schemes, a topic that, while crucial, falls outside the scope of this paper, which is chiefly devoted to advancing semantic segmentation through the introduction of a temporal dimension and specialized attention mechanisms.

\subsection{ISPRS Potsdam Dataset}
\label{valid_sota_isprs}

\begin{figure}
\centering
\includegraphics[clip, trim=0.25cm 0.40cm 0.25cm 0.1cm,width=\columnwidth]{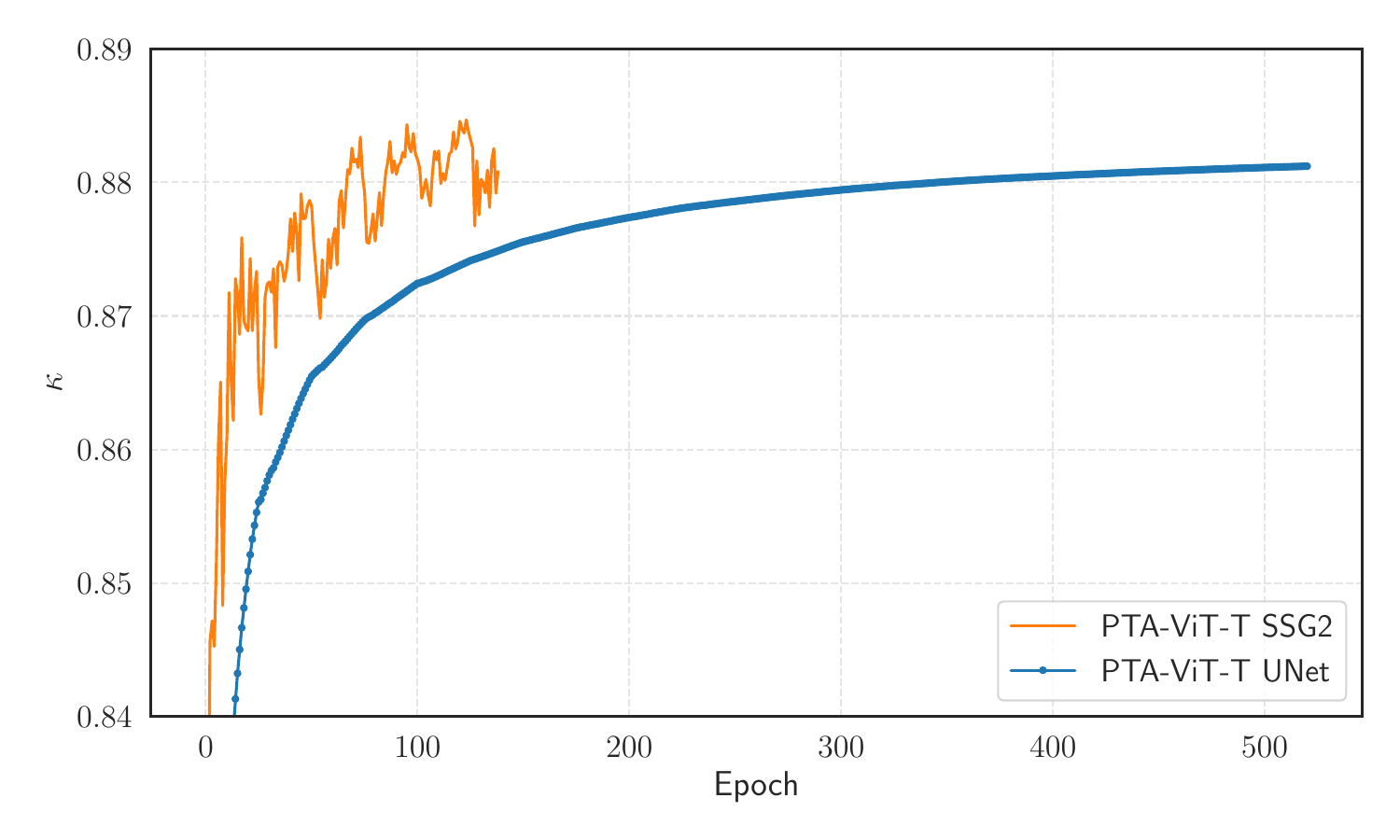}
\caption{Difference in performance evolution, Sequence modelling vs UNet-like modelling on the ISPRS Dataset.}   
  \label{isprs_unet_vs_sequence}
\end{figure}

Our primary aim is to evaluate the efficacy of our SSG2 modeling approach in comparison with well-established methods, particularly focusing on its performance and rate of convergence against UNet-like architectures like PTA-ViT. Table \ref{isprs_results_summary} presents the performance of various state-of-the-art models, replicating the numerical results from \citep[]{Wang_2022}. Top-performing models are highlighted in \textbf{bold}. Both the \texttt{PTA-ViT-T} (UNet-like) and \texttt{SSG2} models exhibit competitive performance with 50M and 75M parameters, respectively. However, it's important to note that the training run for the SSG2 model was terminated prematurely due to computational limitations, as shown in Figure \ref{isprs_unet_vs_sequence}. The model was still in the process of converging at this point.

To further elucidate the convergence advantages, we include results from three separate training runs. The first two runs feature UNet models equipped with \texttt{PTA-ViT} feature extraction units and have an initial embedding dimension of 96 and depths [2,2,5,2]. These models were evaluated at epochs 100 and 521. The third run provides a snapshot of the \texttt{SSG2} model's performance at epoch 117, where it surpasses the UNet models in performance metrics, even after they underwent five times as many gradient updates (epoch 521). This underscores the SSG2 model's superior rate of convergence and overall performance, which is visually demonstrated in Figure \ref{isprs_unet_vs_sequence}. For a thorough evaluation, we also present a stacked inference for the \texttt{SSG2}, averaging predictions from its two best-performing epochs.

\begin{table*}[!ht]
\centering
\footnotesize
\caption{ Here we compare the \texttt{PTA-ViT} SSG2 modelling macro-topology against the UNet macro-topology for the same feature extraction units and different number of data, on the UrbanMonitor dataset (Darwin). The results presented are for $N=16$ support elements for the test  set and for a variable number of support elements for  the validation set.} 
\label{darwin_results_table}
\fontsize{7pt}{8pt}\selectfont
\begin{tabularx}{\linewidth}{l c c c c c c c c } 
\toprule
Model & Num Train Tiles & Num Test Tiles  & epoch &  $\kappa$ (val) & MCC (test) & IoU (test) & $\mathcal{BM}$ (test) & $\mathcal{MK}$ (test) \\
\midrule
PTA-ViT-T 2252nf96 \textbf{F256} - UNet & 1 & 5 & 100 & 89.62 & 93.50 & 97.87 & 92.08 &  \textbf{94.95} \\
PTA-ViT-T 2252nf96 \textbf{F256} - SSG2 (NSupport=2)  & 1 & 5  & 81 & \textbf{ 91.30} & \textbf{93.57} & \textbf{97.89} & \textbf{92.50} & 94.65\\
\hline
PTA-ViT-T 2252nf96 \textbf{F256} - UNet & 4  & 2 & 100 & 93.81 & 94.23 &97.81 & \textbf{94.40} & 94.06  \\
PTA-ViT-T 2252nf96 \textbf{F256} - SSG2 (NSupport=3)   & 4 & 2  & 59 & \textbf{94.42} & \textbf{94.53} & \textbf{97.94} & 94.25 & \textbf{94.80}\\
\hline
\hline
PTA-ViT-T 2252nf96 \textbf{F128} - UNet & 1  & 5 & 100 & 89.22 & 93.49 & 97.86 &  92.81 &  
\textbf{94.17}  \\
PTA-ViT-T 2252nf96 \textbf{F128} - SSG2 (NSupport=12)   & 1 & 5  & 17 & \textbf{90.30} &  \textbf{93.96} &  \textbf{97.99} & \textbf{93.82}  &  94.10\\
\bottomrule
\end{tabularx}
\end{table*}

\begin{figure}[ht!]
\centering
  \includegraphics[width=\linewidth]{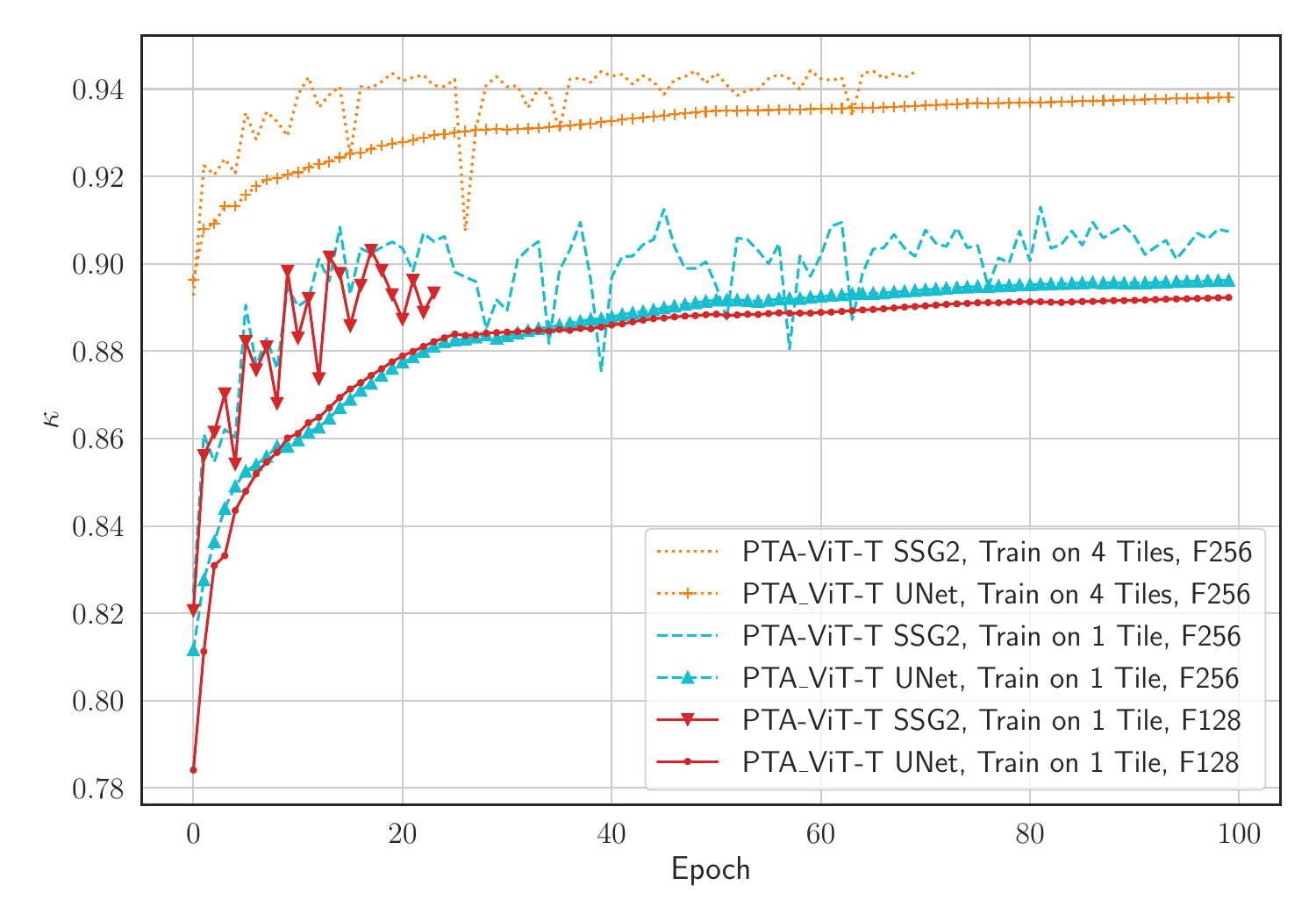}
  \caption{Difference in performance evolution (validation set), for the Darwin Dataset, SSG2 vs UNet-like modelling when using as training images 1 Tile, or 4 Tiles.}
  \label{darwin_unet_vs_sequence}
\end{figure}

\subsection{Performance on varying density datasets}

\label{valid_richpoor_sequence}

Three experiments were designed to assess the performance of UNet-like and Sequence (SSG2) macro-topologies using the \texttt{PTA}-\texttt{ViT}-\texttt{T} baseline model. 
The results are summarized in Table \ref{darwin_results_table} and Figure \ref{darwin_unet_vs_sequence}. 

In the first experiment, we sought to assess the relative performance of the \texttt{SSG2} and \texttt{PTA-ViT-T}  models under the constraints of limited training data and a minimal set of support elements, utilizing a single training tile for this purpose.
The SSG2 model outperformed \texttt{PTA-ViT-T} on several metrics. Specifically, it showed improvements in Cohen's Kappa on the validation set ($\Delta \kappa  = +1.68$), and in  test MCC ($\Delta \text{MCC}=+0.07$) and IoU ($\Delta \text{IoU} = +0.02$). However, these gains were marginal. In terms of Informedness and Markedness, the results were mixed ($\Delta \mathcal{BM} = +0.42$, $\Delta \mathcal{MK} = -0.30$), indicating nuanced differences between the models. Importantly, the distance ($|\mathcal{BM} - \mathcal{MK}|$) was 2.15 for \texttt{SSG2} and 2.87 for the \texttt{PTA-ViT-T}, suggesting that \texttt{SSG2} is a slightly more balanced classifier.

In the third experiment, we used the same dataset but increased the number of support elements for SSG2 to NSupport=12. In comparison with the first experiment, this led to significant gains in SSG2's performance, as evidenced by a \( \Delta \kappa = +1.08 \) in the validation set, and \( \Delta \text{MCC} = +0.47 \) and \( \Delta \text{IoU} = +0.13 \) in the test set. These improvements underscore that the subtle performance differences between SSG2 and \texttt{PTA-ViT-T} in the first experiment were mainly due to the limited number of support elements for SSG2. This point is further emphasized by the near-identical performance of the \texttt{PTA-ViT-T} in MCC and IoU across both experiments. While the metrics for Informedness and Markedness diverged, indicating nuanced differences in qualitative performance, the distance \( | \mathcal{BM} - \mathcal{MK} | \) shrank to 0.28 for SSG2 and 1.36 for \texttt{PTA-ViT-T}. This suggests that both models offer more balanced classifications, although SSG2 maintains a slight edge.

In the second experiment, constrained by GPU memory to a mere three support elements (NSupport=3), SSG2 still managed to outclass \texttt{PTA-ViT-T}. It achieved a \( \Delta \kappa = +0.61 \) on the validation set and \( \Delta \text{MCC} = +0.3 \) and \( \Delta \text{IoU} = +0.13 \) on the test set. Notably, as the dataset size increased, \texttt{PTA-ViT-T} seemed to become more balanced. This is evidenced by a reduced distance \( | \mathcal{BM} - \mathcal{MK} | \) of 0.34, compared to SSG2's 0.55.

To summarize, throughout all experiments, SSG2 consistently outshone UNet-like models, a trend that became more pronounced as the number of support elements increased. This is further supported by the more rapid convergence rates of SSG2, as documented in Figure \ref{darwin_unet_vs_sequence}, making it a more dependable choice for such computational tasks. It is worth noting that the ground truth for UrbanMonitor was predominantly generated using a pre-trained UNet-like model, ResUNet-a  \citep{DIAKOGIANNIS202094}, with some minor manual refinements. This could potentially introduce a bias in the dataset, favoring UNet-like architectures and thereby possibly constraining the scope of what SSG2 can learn beyond the capabilities of UNet-like models.

\subsection{The effect of the number of support elements during inference}
\label{section_effect_of_support_elements}

In this section, we aim to investigate the impact of varying the number of support elements during inference on the model's performance. To this end, we conduct inference tests using 2, 4, 8, 16, and 32 support elements in the sequence set. We employ two models trained on the UrbanMonitor dataset using a single tile, one with NSupport=2 and the other with NSupport=12 (refer to Table \ref{darwin_results_table}).

\begin{figure}[!h]
\centering
\includegraphics[clip, trim=0.05cm 0.40cm 0.25cm 0.1cm,width=\columnwidth]{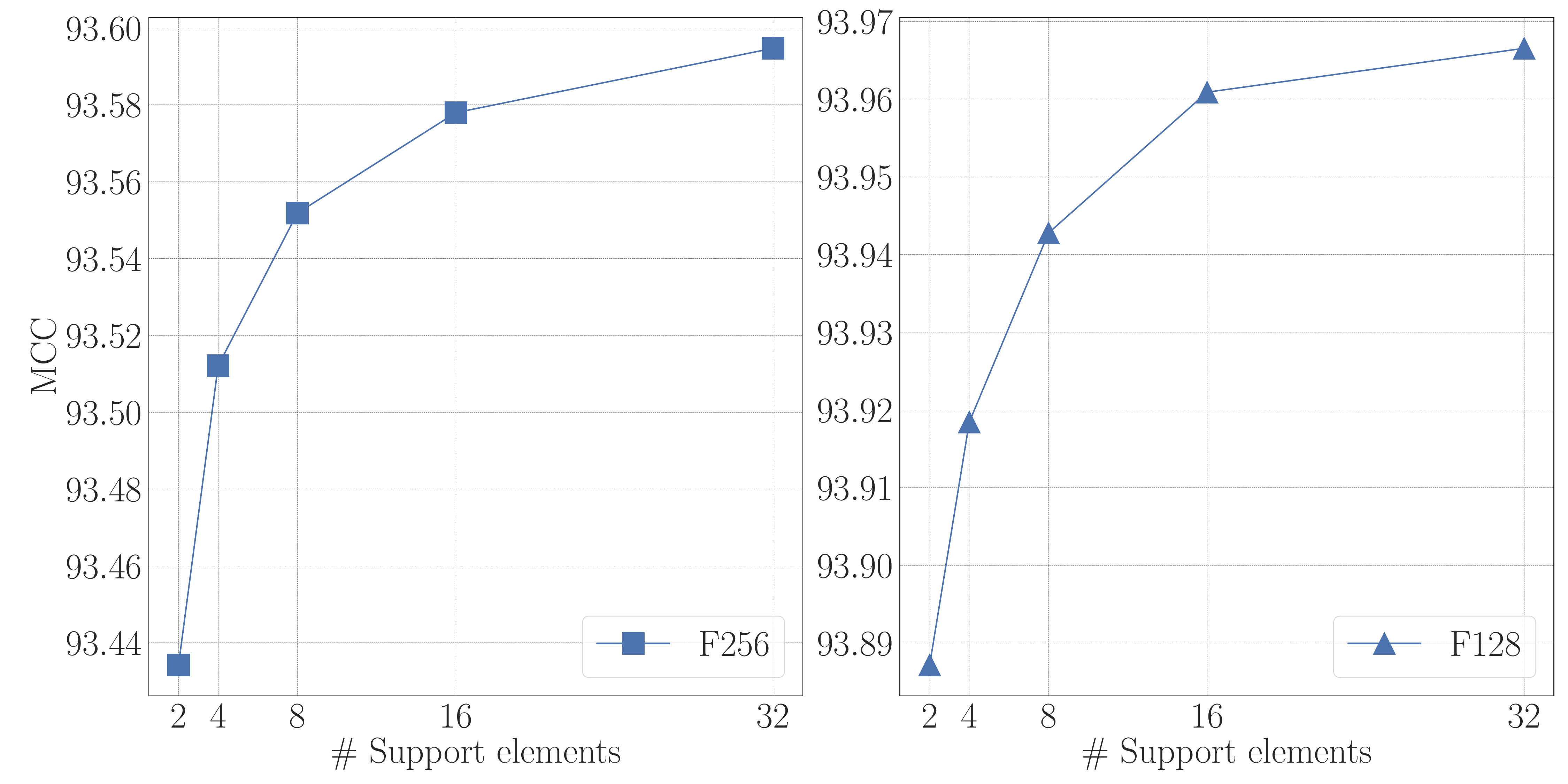}
\caption{Improvement in performance as a function of the number of sequence elements in the support set during inference on the UrbanMonitor dataset (Darwin). Models were trained on a single tile and tested on five tiles. The left panel represents a model trained with NSupport=2, while the right panel depicts a model trained with NSupport=12.
} 
\label{darwin_ablation_mcc}
\end{figure}

Figure \ref{darwin_ablation_mcc} illustrates the change in MCC as we increase the number of support elements during inference. The left panel depicts the model trained with F256 and NSupport=2, while the right panel shows the model trained with F128 and NSupport=12. Both panels indicate that MCC generally improves with an increasing number of support elements, albeit the gains within this range are modest ($\Delta (\text{MCC}_{NS_{\text{train}}=2}) \approx +0.16$,\\  
$\Delta (\text{MCC}_{NS_{\text{train}}=12}) \approx +0.08$).

It's worth noting that the initial MCC varies significantly depending on the NSupport used during training. For instance, when conducting inference with just two support elements, the model trained with NS=12 starts with an MCC of approximately 93.89, whereas the one trained with NS=2 starts at around 93.44, indicating an initial MCC advantage of +0.45 for the former.

Regarding the performance gains achieved by increasing the number of support elements during inference, these continue up to 32 elements, implying potential for further improvements if computational resources permit. This aligns with our initial hypothesis that adding an "extra dimension" to the problem enhances performance, as one would anticipate with multi-observational inference. While the performance improvement shows signs of reaching a plateau, exploring this limit was not feasible due to GPU memory constraints. We expect that future algorithmic optimizations will allow us to delve deeper into this aspect.

\subsection{Learning set operations}
\label{valid_set_operations}
We visualise features from the binary set of buildings that makes easier comprehension of the features learned. 
\begin{figure}[!h]
\centering
\includegraphics[width=\columnwidth]{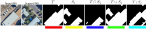}
\includegraphics[clip, trim=0.05cm 0.40cm 0.25cm 0.1cm,width=\columnwidth]{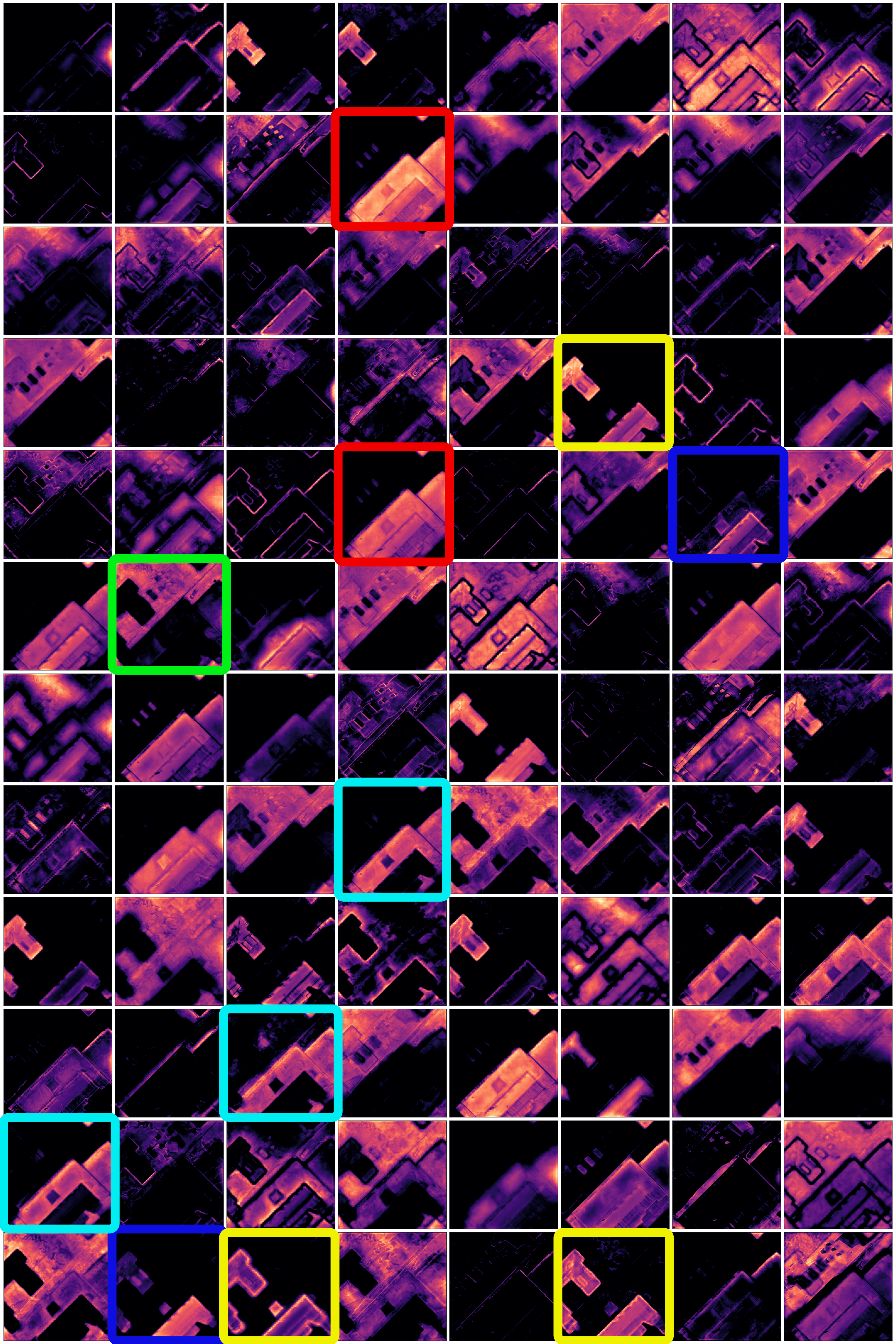}
\caption{Features visualization for the Base feature extractor unit. Color coded squares  correspond to the same color of ground truth masks.} 
\label{features_vis}
\end{figure}
In Fig. \ref{features_vis} we visualize in the first row, from left to right, input target image, support image, their corresponding ground truth masks, and then the set intersection, $T\cap S_j$, the set union, $T\cup S_j$ and the set difference, $T \setminus S_j$. In subsequent rows we visualize all of the final 96 filters of the base feature extractor model that correspond to these predictions (variable \texttt{features} in line 26 of Listing \ref{headcmtsk3d}). We can see that the algorithm learns features that correspond to the ground truth of the support image as well as the ground truth of the target image as well as the set operations (union intersection difference), in addition with features corresponding features that describe boundaries and distance transform.  We note that, for two sets $t_1$ and $t_2$,   
to uniquely reconstruct them, merely knowing their intersection, $t_1 \cap t_2$ and union $t_1 \cup t_2$ is not adequate. Specifically, we also need one of the set differences - e.g. $t_1 \setminus t_2$, to form a complete set of equations that allow for unique determintation of $t_1$ and $t_2$. 

\begin{figure}[!h]
\centering
\includegraphics[clip, trim=0.20cm 1.5cm 0.01cm 1.0cm,width=\columnwidth]{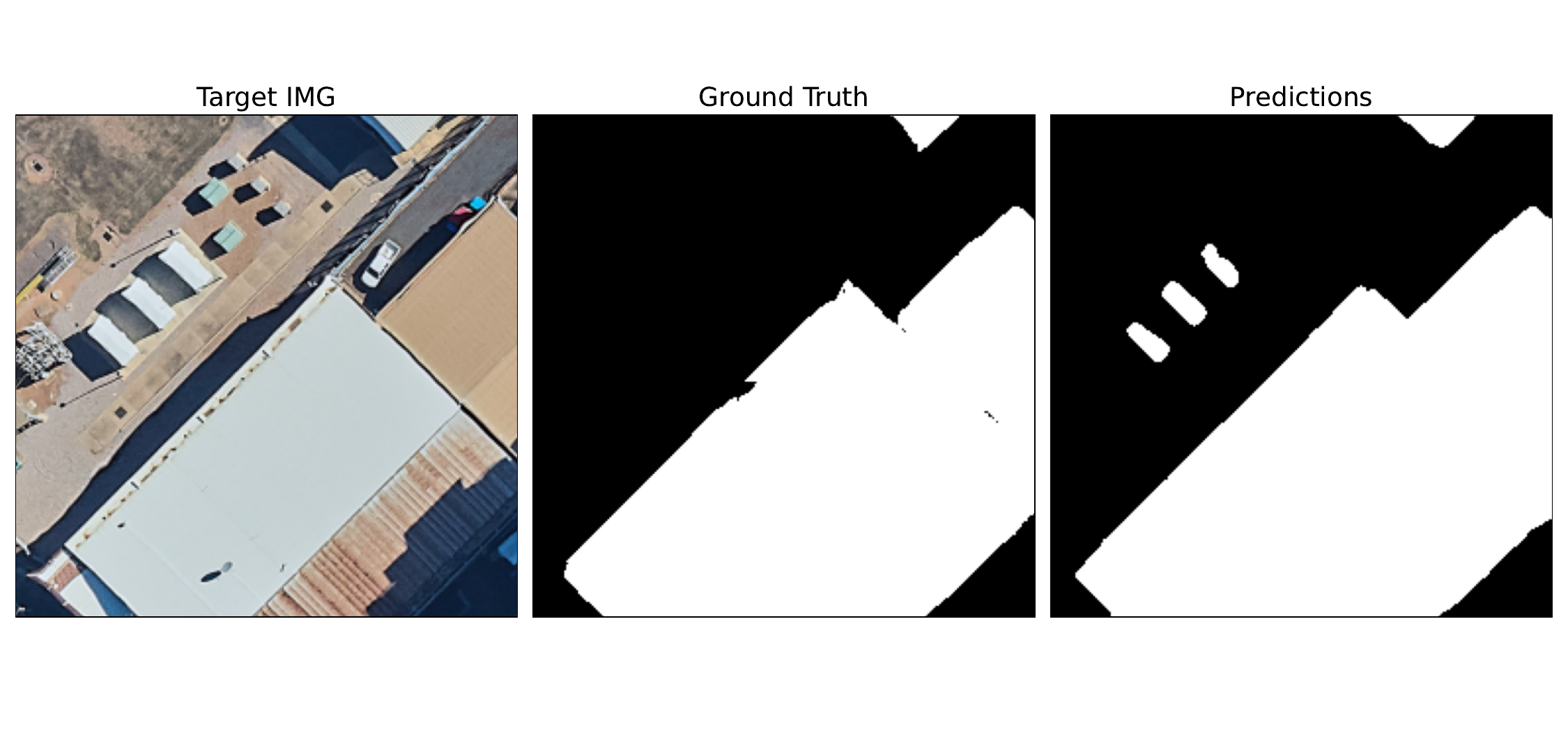}
\includegraphics[clip, trim=0.20cm 1.5cm 0.01cm 1.0cm,width=\columnwidth]{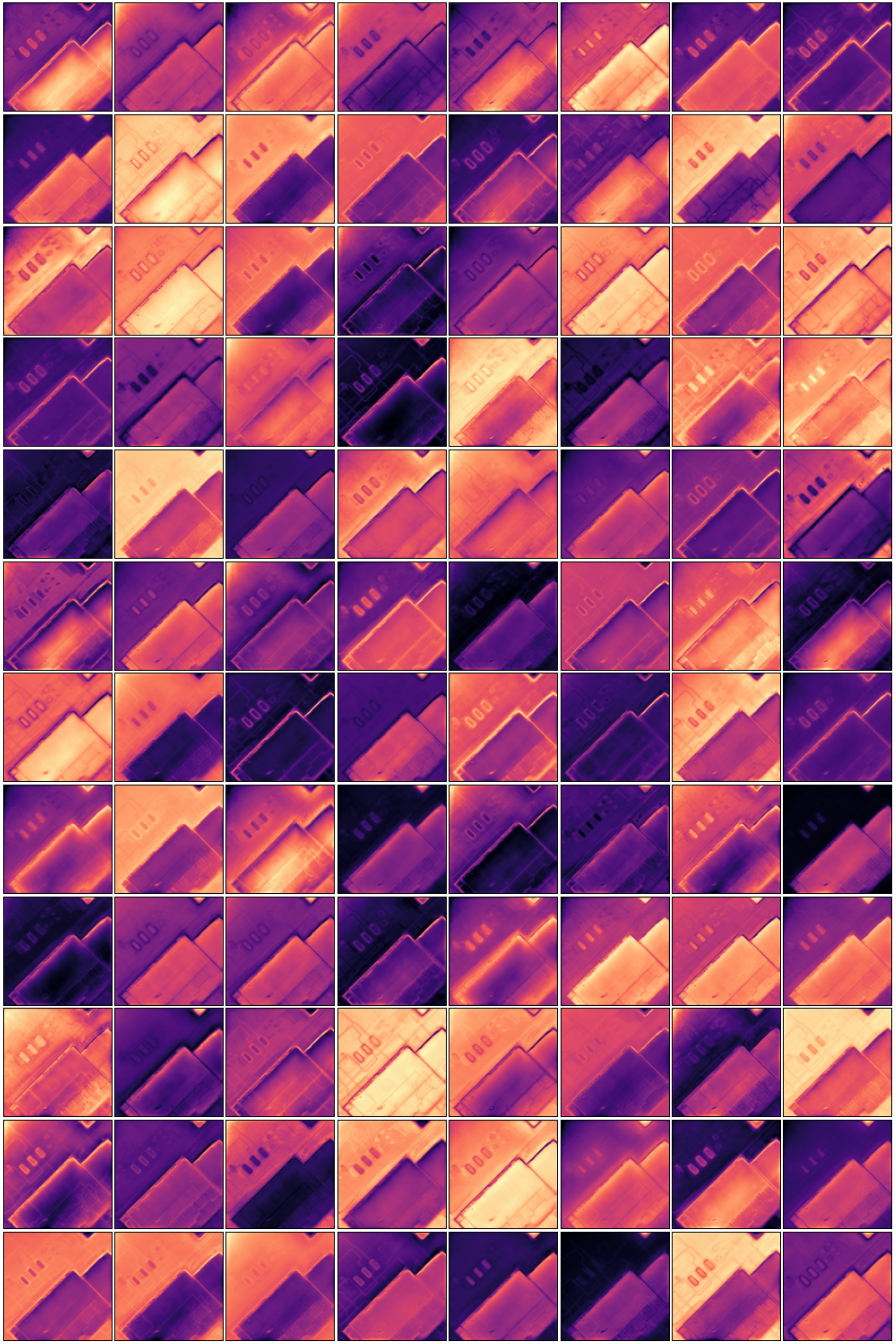}
\caption{Features visualization after applying the 3D feature extraction unit and consuming the prior knowledge of union of intersections as well as intersection of unions. Data are from the Darwin UrbanMonitor dataset} 
\label{features_target}
\end{figure}

In Fig. \ref{features_target} we visualize the  features that are produced after the consumption from the \texttt{PTA-ViT3D} stage (variable \break \texttt{features}\texttt{\_target} on Line 71 of Listing \ref{headcmtsk3d}). We see that these features correspond only to the ground truth of the target image (top row) and all information from support images have been supressed. For example, there is nothing resembling the ground truth of the support image in Fig. \ref{features_vis}. This suggests that the network can indeed filter out and discriminate between target and support images as well as learn from the whole set of features.

\subsection{ISIC 2018 dataset}

\begin{figure}[ht!]
    \centering
    \subfigure[Difference in performance convergence.]{%
    \centering
        \includegraphics[clip, trim=0.25cm 0.40cm 0.25cm 0.1cm,width=\columnwidth]{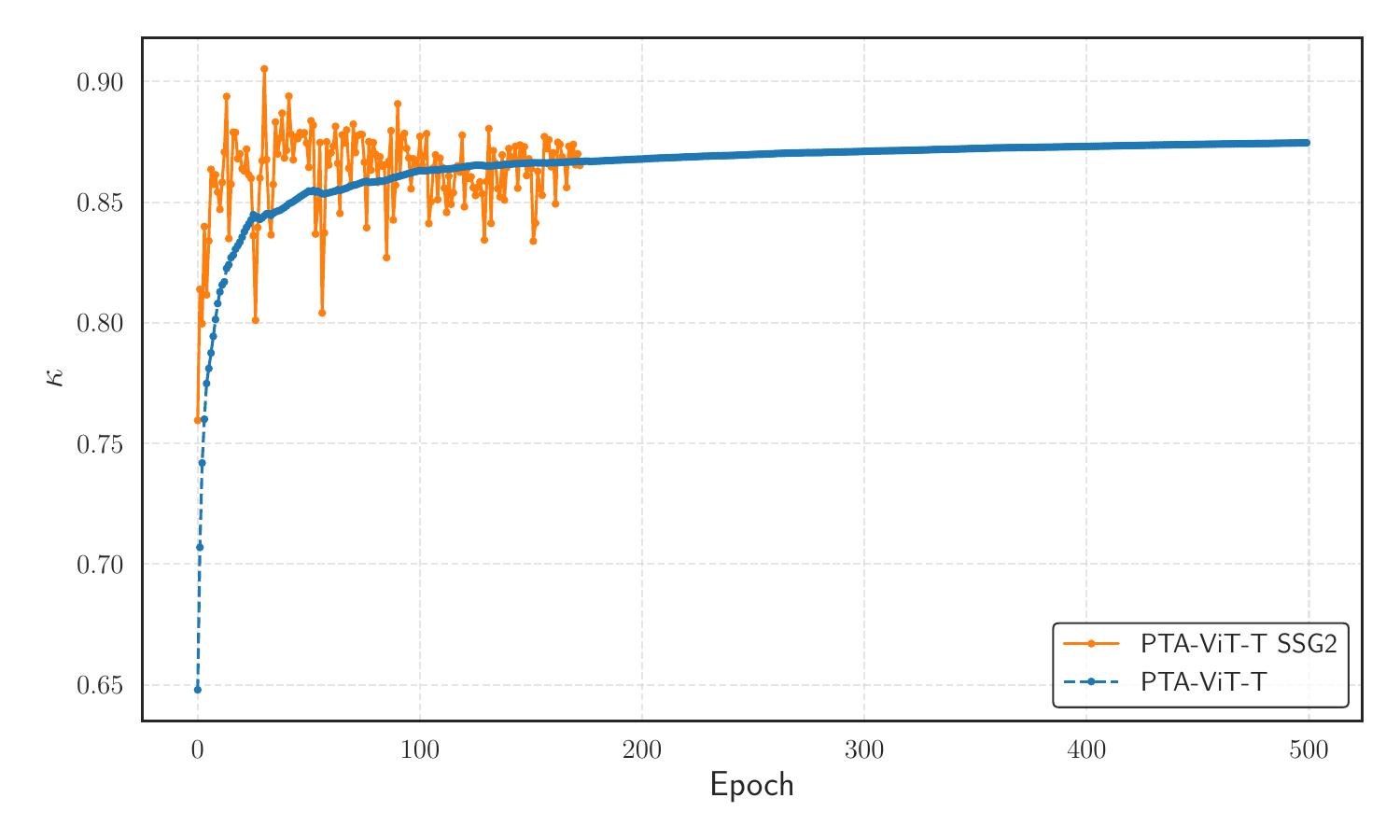}
        \label{isic_2018:suba}
    }
    \subfigure[SSG2 (left) vs UNet modelling, confusion matrices.]{%
        \includegraphics[clip, trim=0.0cm 0.20cm 0.25cm 0.1cm,width=\columnwidth]{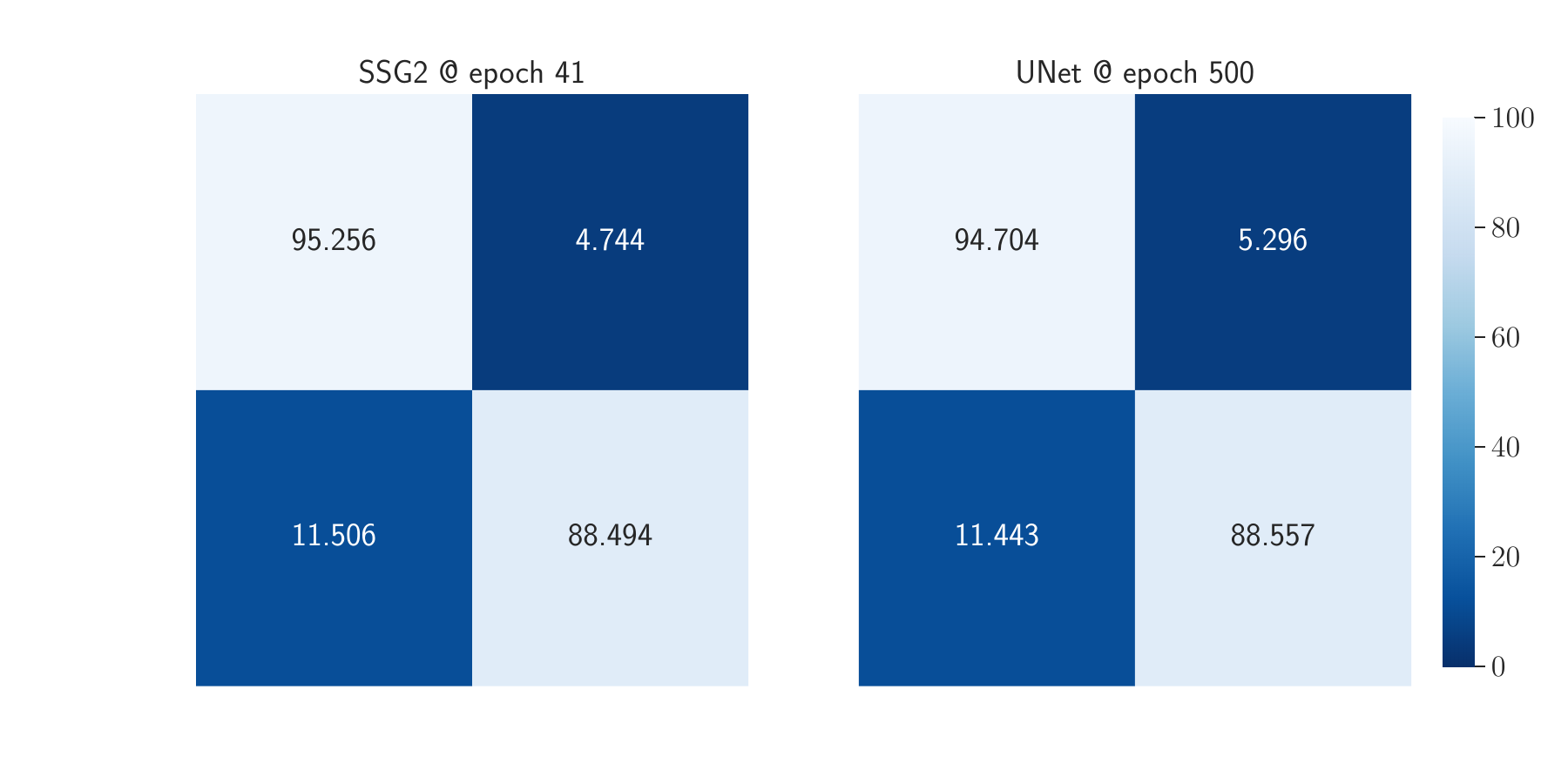}
        \label{isic_2018:subb}
    }
    \caption{Performance difference in convergence and final score between SSG2 modelling and UNet-like on the ISIC2018 Dataset.} 
    \label{isic_2018}
\end{figure}

In an effort to evaluate the generalizability of our SSG2 model -- originally developed for remote sensing applications -- we extend our experiments to the realm of medical imaging, specifically using the ISIC 2018 dataset. This cross-disciplinary test serves to assess how a model tailored for one scientific domain performs when applied to a completely different context, without any modifications.

For the test set, we employ a specific resizing approach. Both the original input and support images are resized to a 256$\times$256 resolution while preserving the aspect ratio. Predicted masks are then scaled back to their original dimensions for comparison with the true segmentation masks. It's important to note that this resizing process inherently limits performance. Specifically, by resizing the ground-truth masks to 256x256 and then back to their original size, the IoU degrades to approximately $\text{IoU} = 95.69694 \pm 6\times 10^-8 \sim 95.7$.  This serves as an upper bound for the IoU that our model, operating at the 256x256 resolution, can realistically achieve.

\begin{table*}[!ht]
\centering
\footnotesize
\caption{ Performance results for the ISIC2018 dataset. The results presented are for $N=16$ support elements for the test  set and for $N=5$ for the validation set. Performance metrics for the Polar ResUNet++, the Boundary Aware Transformer models (BAT) and the Double UNet  are replicated from their corresponding publications.} 
\label{isic2018_results_table}
\begin{tabularx}{\linewidth}{l c c c c c c c} 
\toprule
Model & epoch &  $\kappa$ (val) & MCC (test) & Dice (test) & IoU (test) & $\mathcal{BM}$ (test) & $\mathcal{MK}$ (test) \\
\midrule
Polar Res-U-Net++ \citep{9551998} & 200 & - & - & 92.53 & - & - & -\\
BAT \citep{tang2022duat}  & 500 & - & - & 91.20 & 84.30 & - & - \\
Double UNet \citep{9183321} & 300 & - & - & 89.62 & - & - & -  \\
\hline\hline 
\texttt{PTA-ViT-T}   - UNet & 500 & 87.46 &  82.99& 94.93& 90.36 &83.26  &  82.71 \\
\texttt{PTA-ViT-T}  - SSG2 (NSupport=5)  & 41 & \textbf{89.46} & \textbf{83.80} & \textbf{95.21} & \textbf{90.85}  & \textbf{83.75} & \textbf{83.86}\\
\bottomrule
\end{tabularx}
\end{table*}

In Fig. \ref{isic_2018:suba}, we plot the validation MCC metric as a function of epochs for both the SSG2 and \texttt{PTA}-\texttt{ViT}-\texttt{T} models. Fig. \ref{isic_2018:subb} shows the confusion matrices for these models at epochs 45 and 500. The \texttt{SSG2} model  notably reaches its optimal performance by epoch 45, a milestone the UNet-like architecture fails to meet even with ten times as many gradient updates. Additional metrics, including the Dice coefficient, IoU, Informedness, and Markedness, are presented in Table \ref{isic2018_results_table}, along with documented values of latest published state of the art models \citep{9183321,9551998,tang2022duat}. Across these metrics, the \texttt{SSG2} model consistently outperforms the baseline UNet, indicating its versatility across different applications. 

Remarkably, this performance was attained without any model architecture alterations or data-specific augmentations; the model was directly applied to a new domain.

\section{Discussion}

In the present study, we propose a novel approach to noise reduction in the realm of semantic segmentation via sequence modelling. The underlying premise is that the predicted segmentation mask, which constitutes the signal, exhibits strong correlation across all sequence elements, while noise remains random and uncorrelated. This enables the algorithm to effectively discern between signal and noise. This concept draws parallels to temporal integration in video perception, where the human brain aggregates information over time to smooth out noise and imperfections in individual frames. In essence, sequence modelling in semantic segmentation introduces an additional temporal dimension to the data, thereby facilitating noise suppression.

One noteworthy advantage of our algorithm is its rapid convergence to optimal performance metrics when compared to traditional UNet architectures (Figures \ref{isprs_unet_vs_sequence}, \ref{darwin_unet_vs_sequence}, \ref{isic_2018:suba}). However, this comes at the cost of increased computational demands, manifesting as a higher memory footprint. 
In practice, the maximum batch size used in a UNet framework, needs to be split into a new batch size and a sequence dimension. This poses a limitation on exploring configurations with a greater number of sequence elements but offers avenues for optimization through software enhancements. We hypothesize that future implementations incorporating more sequence elements will yield further improvements in performance.

Embedded within our experiments is an adaptation of the original \texttt{MaxViT} architecture, where we replaced the standard multi-axis attention blocks with our Patch Tanimoto Attention (PTA). This modified version, denoted as \texttt{PTA-ViT}, demonstrates enhanced performance with fewer parameters and a reduced memory footprint (approximately 33\% for the CIFAR10 dataset).
Another incremental  contribution is the inclusion of the \texttt{d2s} activation function  within the attention blocks of our deep networks. Compared to the traditional sigmoid, tanh and \texttt{ReLU} activations, \texttt{d2s} aids in stable convergence, further reinforcing the robustness of our approach.

Regarding the convergence characteristics of the \texttt{SSG2} model, it's important to clarify that our evaluations encompassed experiments with both consistent and varied learning rates and batch sizes for the compared architectures. Across all experimental conditions, \texttt{SSG2} consistently outperformed its counterparts in terms of faster convergence and overall performance, corroborating the results presented herein.

Although our architecture was constructed and tailored for remote sensing aerial data, we show cross discipline competitive performance without any modifications of the modelling framework, to a medical imaging dataset,  the ISIC 2018 skin lesion segmentation. 

Lastly, the versatility of the \texttt{SSG2} framework should not go unmentioned. The architecture is not tightly bound to the specific feature extraction model employed in this study. It is compatible with any deep learning framework designed for similarity or change detection, offering considerable scope for future research and improvements.

\section*{Acknowledgments}
The Authors would like to thank Chen Wu for fruitful conversations and discussions on the methods developed in the manuscript.  
This project was supported by resources and expertise provided by CSIRO IMT Scientific Computing.  
This work was supported by resources provided by The Pawsey Supercomputing Centre with funding from the Australian Government and the Government of Western Australia. The Authors are grateful to the Darwin Living Laboratory and NT Government for access to photography used for the creation of the UrbanMonitor dataset.
The authors acknowledge  the support of the \textsc{pytorch} community.  

\bibliography{AI_BIB}

\begin{thebibliography}{39}
\expandafter\ifx\csname natexlab\endcsname\relax\def\natexlab#1{#1}\fi
\providecommand{\url}[1]{\texttt{#1}}
\providecommand{\href}[2]{#2}
\providecommand{\path}[1]{#1}
\providecommand{\DOIprefix}{doi:}
\providecommand{\ArXivprefix}{arXiv:}
\providecommand{\URLprefix}{URL: }
\providecommand{\Pubmedprefix}{pmid:}
\providecommand{\doi}[1]{\href{http://dx.doi.org/#1}{\path{#1}}}
\providecommand{\Pubmed}[1]{\href{pmid:#1}{\path{#1}}}
\providecommand{\bibinfo}[2]{#2}
\ifx\xfnm\relax \def\xfnm[#1]{\unskip,\space#1}\fi
%Type = Article
\bibitem[{Audebert et~al.(2018)Audebert, {Le Saux} and
  Lefèvre}]{audebert2018beyond}
\bibinfo{author}{Audebert, N.}, \bibinfo{author}{{Le Saux}, B.},
  \bibinfo{author}{Lefèvre, S.}, \bibinfo{year}{2018}.
\newblock \bibinfo{title}{Beyond rgb: Very high resolution urban remote sensing
  with multimodal deep networks}.
\newblock \bibinfo{journal}{ISPRS Journal of Photogrammetry and Remote Sensing}
  \bibinfo{volume}{140}, \bibinfo{pages}{20–32}.
%Type = Article
\bibitem[{Benčević et~al.(2021)Benčević, Galić, Habijan and
  Babin}]{9551998}
\bibinfo{author}{Benčević, M.}, \bibinfo{author}{Galić, I.},
  \bibinfo{author}{Habijan, M.}, \bibinfo{author}{Babin, D.},
  \bibinfo{year}{2021}.
\newblock \bibinfo{title}{Training on polar image transformations improves
  biomedical image segmentation}.
\newblock \bibinfo{journal}{IEEE Access} \bibinfo{volume}{9},
  \bibinfo{pages}{133365–133375}.
\newblock \DOIprefix\doi{10.1109/ACCESS.2021.3116265}.
%Type = Article
\bibitem[{Buslaev et~al.(2020)Buslaev, Iglovikov, Khvedchenya, Parinov,
  Druzhinin and Kalinin}]{info11020125}
\bibinfo{author}{Buslaev, A.}, \bibinfo{author}{Iglovikov, V.I.},
  \bibinfo{author}{Khvedchenya, E.}, \bibinfo{author}{Parinov, A.},
  \bibinfo{author}{Druzhinin, M.}, \bibinfo{author}{Kalinin, A.A.},
  \bibinfo{year}{2020}.
\newblock \bibinfo{title}{Albumentations: Fast and flexible image
  augmentations}.
\newblock \bibinfo{journal}{Information} \bibinfo{volume}{11}.
\newblock \URLprefix \url{https://www.mdpi.com/2078-2489/11/2/125},
  \DOIprefix\doi{10.3390/info11020125}.
%Type = Article
\bibitem[{Codella et~al.(2019)Codella, Rotemberg, Tschandl, Celebi, Dusza,
  Gutman, Helba, Kalloo, Liopyris, Marchetti, Kittler and
  Halpern}]{DBLP:journals/corr/abs-1902-03368}
\bibinfo{author}{Codella, N.C.F.}, \bibinfo{author}{Rotemberg, V.},
  \bibinfo{author}{Tschandl, P.}, \bibinfo{author}{Celebi, M.E.},
  \bibinfo{author}{Dusza, S.W.}, \bibinfo{author}{Gutman, D.A.},
  \bibinfo{author}{Helba, B.}, \bibinfo{author}{Kalloo, A.},
  \bibinfo{author}{Liopyris, K.}, \bibinfo{author}{Marchetti, M.A.},
  \bibinfo{author}{Kittler, H.}, \bibinfo{author}{Halpern, A.},
  \bibinfo{year}{2019}.
\newblock \bibinfo{title}{Skin lesion analysis toward melanoma detection 2018:
  {A} challenge hosted by the international skin imaging collaboration
  {(ISIC)}}.
\newblock \bibinfo{journal}{CoRR} \bibinfo{volume}{abs/1902.03368}.
\newblock \URLprefix \url{http://arxiv.org/abs/1902.03368},
  \href{http://arxiv.org/abs/1902.03368}{\tt arXiv:1902.03368}.
%Type = Article
\bibitem[{Diakogiannis et~al.(2021)Diakogiannis, Waldner and
  Caccetta}]{rs13183707}
\bibinfo{author}{Diakogiannis, F.I.}, \bibinfo{author}{Waldner, F.},
  \bibinfo{author}{Caccetta, P.}, \bibinfo{year}{2021}.
\newblock \bibinfo{title}{Looking for change? roll the dice and demand
  attention}.
\newblock \bibinfo{journal}{Remote Sensing} \bibinfo{volume}{13}.
\newblock \URLprefix \url{https://www.mdpi.com/2072-4292/13/18/3707},
  \DOIprefix\doi{10.3390/rs13183707}.
%Type = Article
\bibitem[{Diakogiannis et~al.(2020)Diakogiannis, Waldner, Caccetta and
  Wu}]{DIAKOGIANNIS202094}
\bibinfo{author}{Diakogiannis, F.I.}, \bibinfo{author}{Waldner, F.},
  \bibinfo{author}{Caccetta, P.}, \bibinfo{author}{Wu, C.},
  \bibinfo{year}{2020}.
\newblock \bibinfo{title}{Resunet-a: A deep learning framework for semantic
  segmentation of remotely sensed data}.
\newblock \bibinfo{journal}{ISPRS Journal of Photogrammetry and Remote Sensing}
  \bibinfo{volume}{162}, \bibinfo{pages}{94–114}.
\newblock \URLprefix
  \url{http://www.sciencedirect.com/science/article/pii/S0924271620300149},
  \DOIprefix\doi{10.1016/j.isprsjprs.2020.01.013}.
%Type = Article
\bibitem[{Dice(1945)}]{doi:10.2307/1932409}
\bibinfo{author}{Dice, L.R.}, \bibinfo{year}{1945}.
\newblock \bibinfo{title}{Measures of the amount of ecologic association
  between species}.
\newblock \bibinfo{journal}{Ecology} \bibinfo{volume}{26},
  \bibinfo{pages}{297–302}.
\newblock \DOIprefix\doi{10.2307/1932409}.
%Type = Article
\bibitem[{Dosovitskiy et~al.(2020)Dosovitskiy, Beyer, Kolesnikov, Weissenborn,
  Zhai, Unterthiner, Dehghani, Minderer, Heigold, Gelly, Uszkoreit and
  Houlsby}]{DBLP:journals/corr/abs-2010-11929}
\bibinfo{author}{Dosovitskiy, A.}, \bibinfo{author}{Beyer, L.},
  \bibinfo{author}{Kolesnikov, A.}, \bibinfo{author}{Weissenborn, D.},
  \bibinfo{author}{Zhai, X.}, \bibinfo{author}{Unterthiner, T.},
  \bibinfo{author}{Dehghani, M.}, \bibinfo{author}{Minderer, M.},
  \bibinfo{author}{Heigold, G.}, \bibinfo{author}{Gelly, S.},
  \bibinfo{author}{Uszkoreit, J.}, \bibinfo{author}{Houlsby, N.},
  \bibinfo{year}{2020}.
\newblock \bibinfo{title}{An image is worth 16x16 words: Transformers for image
  recognition at scale}.
\newblock \bibinfo{journal}{CoRR} \bibinfo{volume}{abs/2010.11929}.
\newblock \URLprefix \url{https://arxiv.org/abs/2010.11929},
  \href{http://arxiv.org/abs/2010.11929}{\tt arXiv:2010.11929}.
%Type = Article
\bibitem[{Gheller et~al.(2023)Gheller, Taffoni and Goz}]{10.1093/rasti/rzad002}
\bibinfo{author}{Gheller, C.}, \bibinfo{author}{Taffoni, G.},
  \bibinfo{author}{Goz, D.}, \bibinfo{year}{2023}.
\newblock \bibinfo{title}{High performance w-stacking for imaging radio
  astronomy data: a parallel and accelerated solution}.
\newblock \bibinfo{journal}{RAS Techniques and Instruments}
  \bibinfo{volume}{2}, \bibinfo{pages}{91–105}.
\newblock \URLprefix \url{https://doi.org/10.1093/rasti/rzad002},
  \DOIprefix\doi{10.1093/rasti/rzad002}.
%Type = Inproceedings
\bibitem[{Glorot and Bengio(2010)}]{pmlr-v9-glorot10a}
\bibinfo{author}{Glorot, X.}, \bibinfo{author}{Bengio, Y.},
  \bibinfo{year}{2010}.
\newblock \bibinfo{title}{Understanding the difficulty of training deep
  feedforward neural networks}, in: \bibinfo{editor}{Teh, Y.W.},
  \bibinfo{editor}{Titterington, M.} (Eds.), \bibinfo{booktitle}{Proceedings of
  the Thirteenth International Conference on Artificial Intelligence and
  Statistics}, \bibinfo{publisher}{PMLR}, \bibinfo{address}{Chia Laguna Resort,
  Sardinia, Italy}. p. \bibinfo{pages}{249–256}.
\newblock \URLprefix \url{https://proceedings.mlr.press/v9/glorot10a.html}.
%Type = Article
\bibitem[{Gorodkin(2004)}]{GORODKIN2004367}
\bibinfo{author}{Gorodkin, J.}, \bibinfo{year}{2004}.
\newblock \bibinfo{title}{Comparing two k-category assignments by a k-category
  correlation coefficient}.
\newblock \bibinfo{journal}{Computational Biology and Chemistry}
  \bibinfo{volume}{28}, \bibinfo{pages}{367–374}.
\newblock \URLprefix
  \url{https://www.sciencedirect.com/science/article/pii/S1476927104000799},
  \DOIprefix\doi{10.1016/j.compbiolchem.2004.09.006}.
%Type = Inproceedings
\bibitem[{Hadsell et~al.(2006)Hadsell, Chopra and
  LeCun}]{c441a21f627f4b0a896fa62cb143176f}
\bibinfo{author}{Hadsell, R.}, \bibinfo{author}{Chopra, S.},
  \bibinfo{author}{LeCun, Y.}, \bibinfo{year}{2006}.
\newblock \bibinfo{title}{Dimensionality reduction by learning an invariant
  mapping}, in: \bibinfo{booktitle}{Proceedings - 2006 IEEE Computer Society
  Conference on Computer Vision and Pattern Recognition, CVPR 2006}, p.
  \bibinfo{pages}{1735–1742}.
\newblock \DOIprefix\doi{10.1109/CVPR.2006.100}. \bibinfo{note}{2006 IEEE
  Computer Society Conference on Computer Vision and Pattern Recognition, CVPR
  2006 ; Conference date: 17-06-2006 Through 22-06-2006}.
%Type = Article
\bibitem[{Haghighi et~al.(2018)Haghighi, Jasemi, Hessabi and
  Zolanvari}]{Haghighi2018-2}
\bibinfo{author}{Haghighi, S.}, \bibinfo{author}{Jasemi, M.},
  \bibinfo{author}{Hessabi, S.}, \bibinfo{author}{Zolanvari, A.},
  \bibinfo{year}{2018}.
\newblock \bibinfo{title}{{PyCM}: Multiclass confusion matrix library in
  python}.
\newblock \bibinfo{journal}{Journal of Open Source Software}
  \bibinfo{volume}{3}, \bibinfo{pages}{729}.
\newblock \URLprefix \url{https://doi.org/10.21105/joss.00729},
  \DOIprefix\doi{10.21105/joss.00729}.
%Type = Inproceedings
\bibitem[{Hamacher(1978)}]{hamacher1978uber}
\bibinfo{author}{Hamacher, H.}, \bibinfo{year}{1978}.
\newblock \bibinfo{title}{Uber logische verknunpfungenn unssharfer aussagen
  undderen zugenhorige bewertungsfunktione}, in: \bibinfo{editor}{Trappl, R.},
  \bibinfo{editor}{Klir, G.J.}, \bibinfo{editor}{Riccardi, A.} (Eds.),
  \bibinfo{booktitle}{Progress in Cybernetics and Systems Research, Vol 3},
  \bibinfo{publisher}{Hemisphere}, \bibinfo{address}{Washington}. p.
  \bibinfo{pages}{276–288}.
%Type = Inproceedings
\bibitem[{He et~al.(2015)He, Zhang, Ren and Sun}]{7410480}
\bibinfo{author}{He, K.}, \bibinfo{author}{Zhang, X.}, \bibinfo{author}{Ren,
  S.}, \bibinfo{author}{Sun, J.}, \bibinfo{year}{2015}.
\newblock \bibinfo{title}{Delving deep into rectifiers: Surpassing human-level
  performance on imagenet classification}, in: \bibinfo{booktitle}{2015 IEEE
  International Conference on Computer Vision (ICCV)}, p.
  \bibinfo{pages}{1026–1034}.
\newblock \DOIprefix\doi{10.1109/ICCV.2015.123}.
%Type = Misc
\bibitem[{{ISPRS} and {BSF Swissphoto}()}]{isprs_bsf_swissphoto_2023}
\bibinfo{author}{{ISPRS}}, \bibinfo{author}{{BSF Swissphoto}}, .
\newblock \bibinfo{title}{Wg3 potsdam overhead data}.
\newblock \URLprefix
  \url{http://www2.isprs.org/commissions/comm3/wg4/tests.html}.
%Type = Inproceedings
\bibitem[{Jha et~al.(2020)Jha, Riegler, Johansen, Halvorsen and
  Johansen}]{9183321}
\bibinfo{author}{Jha, D.}, \bibinfo{author}{Riegler, M.A.},
  \bibinfo{author}{Johansen, D.}, \bibinfo{author}{Halvorsen, P.},
  \bibinfo{author}{Johansen, H.D.}, \bibinfo{year}{2020}.
\newblock \bibinfo{title}{Doubleu-net: A deep convolutional neural network for
  medical image segmentation}, in: \bibinfo{booktitle}{2020 IEEE 33rd
  International Symposium on Computer-Based Medical Systems (CBMS)}, p.
  \bibinfo{pages}{558–564}.
\newblock \DOIprefix\doi{10.1109/CBMS49503.2020.00111}.
%Type = Inproceedings
\bibitem[{Koch et~al.(2015)Koch, Zemel and Salakhutdinov}]{Koch2015SiameseNN}
\bibinfo{author}{Koch, G.}, \bibinfo{author}{Zemel, R.},
  \bibinfo{author}{Salakhutdinov, R.}, \bibinfo{year}{2015}.
\newblock \bibinfo{title}{Siamese neural networks for one-shot image
  recognition}.
%Type = Article
\bibitem[{Kurczynski and Gawiser(2010)}]{Kurczynski_2010}
\bibinfo{author}{Kurczynski, P.}, \bibinfo{author}{Gawiser, E.},
  \bibinfo{year}{2010}.
\newblock \bibinfo{title}{A simultaneous stacking and deblending algorithm for
  astronomical images}.
\newblock \bibinfo{journal}{The Astronomical Journal} \bibinfo{volume}{139},
  \bibinfo{pages}{1592}.
\newblock \URLprefix \url{https://dx.doi.org/10.1088/0004-6256/139/4/1592},
  \DOIprefix\doi{10.1088/0004-6256/139/4/1592}.
%Type = Article
\bibitem[{Litjens et~al.(2017)Litjens, Kooi, Bejnordi, Setio, Ciompi,
  Ghafoorian, {van der Laak}, {van Ginneken} and Sánchez}]{LITJENS201760}
\bibinfo{author}{Litjens, G.}, \bibinfo{author}{Kooi, T.},
  \bibinfo{author}{Bejnordi, B.E.}, \bibinfo{author}{Setio, A.A.A.},
  \bibinfo{author}{Ciompi, F.}, \bibinfo{author}{Ghafoorian, M.},
  \bibinfo{author}{{van der Laak}, J.A.}, \bibinfo{author}{{van Ginneken}, B.},
  \bibinfo{author}{Sánchez, C.I.}, \bibinfo{year}{2017}.
\newblock \bibinfo{title}{A survey on deep learning in medical image analysis}.
\newblock \bibinfo{journal}{Medical Image Analysis} \bibinfo{volume}{42},
  \bibinfo{pages}{60–88}.
\newblock \URLprefix
  \url{https://www.sciencedirect.com/science/article/pii/S1361841517301135},
  \DOIprefix\doi{10.1016/j.media.2017.07.005}.
%Type = Inproceedings
\bibitem[{Liu et~al.(2020)Liu, Jiang, He, Chen, Liu, Gao and Han}]{Liu2020On}
\bibinfo{author}{Liu, L.}, \bibinfo{author}{Jiang, H.}, \bibinfo{author}{He,
  P.}, \bibinfo{author}{Chen, W.}, \bibinfo{author}{Liu, X.},
  \bibinfo{author}{Gao, J.}, \bibinfo{author}{Han, J.}, \bibinfo{year}{2020}.
\newblock \bibinfo{title}{On the variance of the adaptive learning rate and
  beyond}, in: \bibinfo{booktitle}{International Conference on Learning
  Representations}.
\newblock \URLprefix \url{https://openreview.net/forum?id=rkgz2aEKDr}.
%Type = Inproceedings
\bibitem[{Loshchilov and Hutter(2017)}]{loshchilov2016sgdr}
\bibinfo{author}{Loshchilov, I.}, \bibinfo{author}{Hutter, F.},
  \bibinfo{year}{2017}.
\newblock \bibinfo{title}{Sgdr: Stochastic gradient descent with warm
  restarts}, in: \bibinfo{booktitle}{Proceedings of the 5th International
  Conference on Learning Representations (ICLR)}.
\newblock \URLprefix \url{https://openreview.net/forum?id=Skq89Scxx}.
%Type = Article
\bibitem[{Matthews(1975)}]{MATTHEWS1975442}
\bibinfo{author}{Matthews, B.}, \bibinfo{year}{1975}.
\newblock \bibinfo{title}{Comparison of the predicted and observed secondary
  structure of t4 phage lysozyme}.
\newblock \bibinfo{journal}{Biochimica et Biophysica Acta (BBA) - Protein
  Structure} \bibinfo{volume}{405}, \bibinfo{pages}{442–451}.
\newblock \URLprefix
  \url{http://www.sciencedirect.com/science/article/pii/0005279575901099},
  \DOIprefix\doi{10.1016/0005-2795(75)90109-9}.
%Type = Article
\bibitem[{Mo et~al.(2022)Mo, Wu, Yang, Liu and Liao}]{MO2022626}
\bibinfo{author}{Mo, Y.}, \bibinfo{author}{Wu, Y.}, \bibinfo{author}{Yang, X.},
  \bibinfo{author}{Liu, F.}, \bibinfo{author}{Liao, Y.}, \bibinfo{year}{2022}.
\newblock \bibinfo{title}{Review the state-of-the-art technologies of semantic
  segmentation based on deep learning}.
\newblock \bibinfo{journal}{Neurocomputing} \bibinfo{volume}{493},
  \bibinfo{pages}{626–646}.
\newblock \URLprefix
  \url{https://www.sciencedirect.com/science/article/pii/S0925231222000054},
  \DOIprefix\doi{10.1016/j.neucom.2022.01.005}.
%Type = Article
\bibitem[{Nogueira et~al.(2019)Nogueira, {Dalla Mura}, Chanussot, Schwartz and
  dos Santos}]{8727958}
\bibinfo{author}{Nogueira, K.}, \bibinfo{author}{{Dalla Mura}, M.},
  \bibinfo{author}{Chanussot, J.}, \bibinfo{author}{Schwartz, W.R.},
  \bibinfo{author}{dos Santos, J.A.}, \bibinfo{year}{2019}.
\newblock \bibinfo{title}{Dynamic multicontext segmentation of remote sensing
  images based on convolutional networks}.
\newblock \bibinfo{journal}{IEEE Transactions on Geoscience and Remote Sensing}
  \bibinfo{volume}{57}, \bibinfo{pages}{7503–7520}.
\newblock \DOIprefix\doi{10.1109/TGRS.2019.2913861}.
%Type = Article
\bibitem[{Powers(2011)}]{6b87510ce7324df69116f8395644ed77}
\bibinfo{author}{Powers, D.}, \bibinfo{year}{2011}.
\newblock \bibinfo{title}{Evaluation: From precision, recall and f-measure to
  roc, informedness, markedness \& correlation}.
\newblock \bibinfo{journal}{Journal of Machine Learning Technologies}
  \bibinfo{volume}{2}, \bibinfo{pages}{37–63}.
%Type = Article
\bibitem[{Ronneberger et~al.(2015)Ronneberger, Fischer and
  Brox}]{DBLP:journals/corr/RonnebergerFB15}
\bibinfo{author}{Ronneberger, O.}, \bibinfo{author}{Fischer, P.},
  \bibinfo{author}{Brox, T.}, \bibinfo{year}{2015}.
\newblock \bibinfo{title}{U-net: Convolutional networks for biomedical image
  segmentation}.
\newblock \bibinfo{journal}{CoRR} \bibinfo{volume}{abs/1505.04597}.
\newblock \URLprefix \url{http://arxiv.org/abs/1505.04597},
  \href{http://arxiv.org/abs/1505.04597}{\tt arXiv:1505.04597}.
%Type = Article
\bibitem[{Sherrah(2016)}]{Sherrah2016}
\bibinfo{author}{Sherrah, J.}, \bibinfo{year}{2016}.
\newblock \bibinfo{title}{Fully convolutional networks for dense semantic
  labelling of high-resolution aerial imagery}.
\newblock \bibinfo{journal}{CoRR} \bibinfo{volume}{abs/1606.02585}.
\newblock \URLprefix \url{http://arxiv.org/abs/1606.02585},
  \href{http://arxiv.org/abs/1606.02585}{\tt arXiv:1606.02585}.
%Type = Article
\bibitem[{Siddique et~al.(2021)Siddique, Paheding, Elkin and
  Devabhaktuni}]{9446143}
\bibinfo{author}{Siddique, N.}, \bibinfo{author}{Paheding, S.},
  \bibinfo{author}{Elkin, C.P.}, \bibinfo{author}{Devabhaktuni, V.},
  \bibinfo{year}{2021}.
\newblock \bibinfo{title}{U-net and its variants for medical image
  segmentation: A review of theory and applications}.
\newblock \bibinfo{journal}{IEEE Access} \bibinfo{volume}{9},
  \bibinfo{pages}{82031–82057}.
\newblock \DOIprefix\doi{10.1109/ACCESS.2021.3086020}.
%Type = Article
\bibitem[{Sijbers et~al.(1996)Sijbers, Scheunders, Bonnet, {Van Dyck} and
  Raman}]{SIJBERS19961157}
\bibinfo{author}{Sijbers, J.}, \bibinfo{author}{Scheunders, P.},
  \bibinfo{author}{Bonnet, N.}, \bibinfo{author}{{Van Dyck}, D.},
  \bibinfo{author}{Raman, E.}, \bibinfo{year}{1996}.
\newblock \bibinfo{title}{Quantification and improvement of the signal-to-noise
  ratio in a magnetic resonance image acquisition procedure}.
\newblock \bibinfo{journal}{Magnetic Resonance Imaging} \bibinfo{volume}{14},
  \bibinfo{pages}{1157–1163}.
\newblock \URLprefix
  \url{https://www.sciencedirect.com/science/article/pii/S0730725X96002196},
  \DOIprefix\doi{10.1016/S0730-725X(96)00219-6}.
%Type = Inproceedings
\bibitem[{Sutskever et~al.(2013)Sutskever, Martens, Dahl and
  Hinton}]{pmlr-v28-sutskever13}
\bibinfo{author}{Sutskever, I.}, \bibinfo{author}{Martens, J.},
  \bibinfo{author}{Dahl, G.}, \bibinfo{author}{Hinton, G.},
  \bibinfo{year}{2013}.
\newblock \bibinfo{title}{On the importance of initialization and momentum in
  deep learning}, in: \bibinfo{editor}{Dasgupta, S.},
  \bibinfo{editor}{McAllester, D.} (Eds.), \bibinfo{booktitle}{Proceedings of
  the 30th International Conference on Machine Learning},
  \bibinfo{publisher}{PMLR}, \bibinfo{address}{Atlanta, Georgia, USA}. p.
  \bibinfo{pages}{1139–1147}.
\newblock \URLprefix \url{https://proceedings.mlr.press/v28/sutskever13.html}.
%Type = Article
\bibitem[{Taghanaki et~al.(2019)Taghanaki, Abhishek, Cohen, Cohen-Adad and
  Hamarneh}]{taghanaki2019deep}
\bibinfo{author}{Taghanaki, S.A.}, \bibinfo{author}{Abhishek, K.},
  \bibinfo{author}{Cohen, J.P.}, \bibinfo{author}{Cohen-Adad, J.},
  \bibinfo{author}{Hamarneh, G.}, \bibinfo{year}{2019}.
\newblock \bibinfo{title}{Deep semantic segmentation of natural and medical
  images: A review} \href{http://arxiv.org/abs/1910.07655}{\tt
  arXiv:1910.07655}.
%Type = Article
\bibitem[{Tang et~al.(2022)Tang, Huang, Wang, Hou, Su and Liu}]{tang2022duat}
\bibinfo{author}{Tang, F.}, \bibinfo{author}{Huang, Q.}, \bibinfo{author}{Wang,
  J.}, \bibinfo{author}{Hou, X.}, \bibinfo{author}{Su, J.},
  \bibinfo{author}{Liu, J.}, \bibinfo{year}{2022}.
\newblock \bibinfo{title}{Duat: Dual-aggregation transformer network for
  medical image segmentation}.
\newblock \bibinfo{journal}{arXiv preprint arXiv:2212.11677} .
%Type = Article
\bibitem[{Trockman and Kolter(2023)}]{trockman2023patches}
\bibinfo{author}{Trockman, A.}, \bibinfo{author}{Kolter, J.Z.},
  \bibinfo{year}{2023}.
\newblock \bibinfo{title}{Patches are all you need?}
\newblock \bibinfo{journal}{Transactions on Machine Learning Research}
  \URLprefix \url{https://openreview.net/forum?id=rAnB7JSMXL}.
  \bibinfo{note}{featured Certification}.
%Type = Article
\bibitem[{Tschandl et~al.(2018)Tschandl, Rosendahl and
  Kittler}]{tschandl2018ham10000}
\bibinfo{author}{Tschandl, P.}, \bibinfo{author}{Rosendahl, C.},
  \bibinfo{author}{Kittler, H.}, \bibinfo{year}{2018}.
\newblock \bibinfo{title}{The ham10000 dataset, a large collection of
  multi-source dermatoscopic images of common pigmented skin lesions}.
\newblock \bibinfo{journal}{Scientific Data} \bibinfo{volume}{5},
  \bibinfo{pages}{180161}.
\newblock \DOIprefix\doi{10.1038/sdata.2018.161}.
%Type = Article
\bibitem[{Tu et~al.(2022)Tu, Talebi, Zhang, Yang, Milanfar, Bovik and
  Li}]{tu2022maxvit}
\bibinfo{author}{Tu, Z.}, \bibinfo{author}{Talebi, H.}, \bibinfo{author}{Zhang,
  H.}, \bibinfo{author}{Yang, F.}, \bibinfo{author}{Milanfar, P.},
  \bibinfo{author}{Bovik, A.}, \bibinfo{author}{Li, Y.}, \bibinfo{year}{2022}.
\newblock \bibinfo{title}{Maxvit: Multi-axis vision transformer}.
\newblock \bibinfo{journal}{ECCV} .
%Type = Article
\bibitem[{Wang et~al.(2022)Wang, Li, Zhang, Fang, Duan, Meng and
  Atkinson}]{Wang_2022}
\bibinfo{author}{Wang, L.}, \bibinfo{author}{Li, R.}, \bibinfo{author}{Zhang,
  C.}, \bibinfo{author}{Fang, S.}, \bibinfo{author}{Duan, C.},
  \bibinfo{author}{Meng, X.}, \bibinfo{author}{Atkinson, P.M.},
  \bibinfo{year}{2022}.
\newblock \bibinfo{title}{{UNetFormer}: A {UNet}-like transformer for efficient
  semantic segmentation of remote sensing urban scene imagery}.
\newblock \bibinfo{journal}{{ISPRS} Journal of Photogrammetry and Remote
  Sensing} \bibinfo{volume}{190}, \bibinfo{pages}{196–214}.
\newblock \URLprefix \url{https://doi.org/10.1016%2Fj.isprsjprs.2022.06.008},
  \DOIprefix\doi{10.1016/j.isprsjprs.2022.06.008}.
%Type = Article
\bibitem[{Weng(2023)}]{weng2023transformer}
\bibinfo{author}{Weng, L.}, \bibinfo{year}{2023}.
\newblock \bibinfo{title}{The transformer family version 2.0}.
\newblock \bibinfo{journal}{lilianweng.github.io} \URLprefix
  \url{https://lilianweng.github.io/posts/2023-01-27-the-transformer-family-v2/}.
%Type = Article
\bibitem[{{Zhu} et~al.(2017){Zhu}, {Tuia}, {Mou}, {Xia}, {Zhang}, {Xu} and
  {Fraundorfer}}]{8113128}
\bibinfo{author}{{Zhu}, X.X.}, \bibinfo{author}{{Tuia}, D.},
  \bibinfo{author}{{Mou}, L.}, \bibinfo{author}{{Xia}, G.},
  \bibinfo{author}{{Zhang}, L.}, \bibinfo{author}{{Xu}, F.},
  \bibinfo{author}{{Fraundorfer}, F.}, \bibinfo{year}{2017}.
\newblock \bibinfo{title}{Deep learning in remote sensing: A comprehensive
  review and list of resources}.
\newblock \bibinfo{journal}{IEEE Geoscience and Remote Sensing Magazine}
  \bibinfo{volume}{5}, \bibinfo{pages}{8–36}.
\newblock \DOIprefix\doi{10.1109/MGRS.2017.2762307}.

\end{thebibliography}

\appendix

\section{Algorithms}
\label{section_algorithms}
Here we present with \textsc{pytorch} style pseudocode the implementation of some critical components of the modules we developed.

\subsection{Patched Tanimoto Similarity}

\begin{python}[emphstyle=\textcolor{magenta}, caption={\textsc{pytorch} style pseudo code for the $\mathcal{T}(q,k)$ similarity (Eq. \ref{Tanimoto_similarity}).}, emph={qk_sim,return},label={qk_similarity}]
import torch.einsum as einsum
def qk_sim(q,k,smooth=1.e-5):
    # q.shape -> B x (c x h x w) x (C/c x H/h x W/w)
    # k.shape -> B x [c x h x w] x (C/c x H/h x W/w)

    #B x (c x h x w) x [c x h x w]
    qk = einsum('bjklmno,bstrmno->bjklstr',q,k) 
    #B x (c x h x w) 
    qq = einsum('bjklmno,bjklmno->bjkl',q,q) 
    #B x [c x h x w]    
    kk = einsum('bstrmno,bstrmno->bstr',k,k) 

    #B x (c x h x w) x [c x h x w]
    denum = (qq[:,:,:,:,None,None,None]
                 + kk[:,None,None,None])-qk +smooth
    return (qk+smooth)/denum
\end{python}

\subsection{Patch Tanimoto Attention}

The \texttt{PatchifyCHW} function was inspired from this \href{https://discuss.pytorch.org/t/creating-nonoverlapping-patches-from-3d-data-and-reshape-them-back-to-the-image/51210/6?u=foivos_diakogiannis}{discussion} in pytorch community.
\begin{python}[caption={\textsc{pytorch} style pseudocode for splitting a tensor to non overlapping tiles},
emph={PatchifyCHW,_2patch,_2tensor,__init__,  nn,torch,Module},
emphstyle=\textcolor{magenta},label={PatchifyCHW}]
class PatchifyCHW(torch.nn.Module):
    def __init__(self, cscale, hscale, wscale):
        super().__init__()
        self.c = cscale
        self.h = hscale
        self.w = wscale
        self.unfold_shape = None

    def _2patch(self,input):
        shape = input.shape
        # partitions
        c     = shape[1]//self.c
        h     = shape[2]//self.h
        w     = shape[3]//self.w

        # strides
        sc    = c
        sh    = h 
        sw    = w 

        patch = input.unfold(1,c,sc)
        patch = patch.unfold(2,h,sh)
        patch = patch.unfold(3,w,sw)
        
        self.unfold_shape = patch.shape
        
        return patch

    def _2tensor(self, patch):
        B,c1,h1,w1,c2,h2,w2 = self.unfold_shape
        C    = c1 * c2
        H    = h1 * h2 
        W    = w1 * w2
        _tensor = patch.permute(0, 1, 4, 2, 5, 3, 6)
        _tensor = _tensor.view(B,C,H,W)
        return _tensor
\end{python}

\begin{python}[caption={\textsc{pytorch} style pseudocode for patch attention module},
emph={RelPatchAttention2D,forward,__init__,qk_sim, nn,torch,Module},emphstyle=\textcolor{magenta},label={PTAttentionCODE},numbers=left]
import torch.nn as nn
class RelPatchAttention2D(nn.Module):
    def __init__(self, nchannels, nheads, scales):
        super().__init__()
        self.q = Conv2DN(nchannels,groups=nheads)
        self.k = Conv2DN(nchannels,groups=nheads)
        self.v = Conv2DN(nchannels,groups=nheads)
        scales = (c,h,w)
        self.patchify = PatchifyCHW(c,h,w)

        self.shrink2_1 = torch.nn.Linear(c*h*w,1)        
        self.d2s = D2Sigmoid()
        
    def forward(self, input1,input2):
        # query, key, value
        q = self.d2s(self.q(input1))#dim:(B,C,H,W)
        k = self.d2s(self.k(input2))#dim:(B,C,H,W) 
        v = self.v(input2)#dim:(B,C,H,W)

        #dim:(B,c,h,w,C//c,H//h,W//w)
        qp = patchify._2patch(q)  
        kp = patchify._2patch(k)
        vp = patchify._2patch(v)

        # B,(c,h,w),[c,h,w]
        b,c,h,w,_,_,_ = qp.shape
        qksim = qk_sim(qp,kp) 
        qksim = qksim.reshape(qb,-1,c,h,w)
        
        # Sum over the q indices
        # correlate partitions
        qksim = qksim.transpose([0,2,3,4,-1]) 
        qksim = self.shrink2_1(qksim).squeeze()
        
        #dim:(B,c,h,w,C//c,H//h,W//w)
        # element-wise multiplication
        att = qksim * vp
        #dim:(B,C,H,W)
        att = self.patchify._2tensor(att)
        
        return self.d2s(att) 
\end{python}

\subsection{Sequence modelling Algorithm}

\begin{python}[caption={\textsc{pytorch} style pseudocode for SSG2 HEAD. Each of the \texttt{head\_cmtsk} modules creates predictions for boundaries, distance transform and segmentation masks in accordance with \citet{rs13183707}.},
emph={head_cmtsk3D,forward,__init__,fz_inter,fz_union,None, nn,torch,Module},emphstyle=\textcolor{magenta},label={headcmtsk3d},numbers=left]
import torch.nn as nn
class head_cmtsk3D(nn.Module):
    def __init__(self, NClasses, ...):
        super().__init__()
    
        self.NClasses = NClasses 
        self.head_inters = head_cmtsk(NClasses,...)
        self.head_unions = head_cmtsk(NClasses,...)
        self.head_diffs  = head_cmtsk(NClasses,...)
        
        self.compress    = nn.Conv2d(...,k=1)
        self.head_target = head_cmtsk(NClasses,...)
        self.ptavit3d = PTA-ViT3D-Stage(...)        
        
    def forward(self, lst_of_features):
        b,c,s,h,w = lst_of_features.shape
        
        preds_inter          = []
        preds_union          = []
        preds_diffs          = []        
        preds_target_fz      = []
        preds_null_fz        = []

        #for features in lst_of_features:
        for seq_idx in range(s):
            features = lst_of_features[:,:,seq_idx]
            # all set operations result in from 
            # common features of target-support pairs             
            inter =  self.head_inters(features)
            preds_inter.append(inter[:,:,None]) 
            
            union =  self.head_unions(features)
            preds_union.append(union[:,:,None])
     
            diff = self.head_diffs(features)
            preds_diffs.append(diff[:,:,None])

            # inter \cup diff is Target 
            # Here disregard distance/boundaries
            # inter.shape = B,3*NClasses,H,W
            # 3*NClasses=segmenation+bounds+distance
            inter = inter[:,self.NClasses]
            diff  = diff[:,self.NClasses]
            target_fz=fz_conorm(inter,diff)
            preds_target_fz.append(target_fz[:,:,None])

            # inter \cap diff is null 
            # Here disregard distance/boundaries
            null_fz = fz_tnorm(inter,diff)
            preds_null_fz.append(null_fz[:,:,None])
        
        preds_inter      = torch.cat(preds_inter,2)
        preds_union      = torch.cat(preds_union,2)
        preds_diffs      = torch.cat(preds_diffs,2)
        
        # maps to TARGET
        preds_target_fz= torch.cat(preds_target_fz,2) 
        # maps to NULL (zeros) 
        preds_null_fz  = torch.cat(preds_null_fz,2)
        
        # Union of all intersections PRIOR
        uint = torch.max(outs_inter,2)[0] 
        # Intersections of all unions PRIOR
        iuni = torch.min(outs_union,2)[0] 
        
        # This utilizes cross sequence
        # and local spatial correlation
        out3d = self.ptavit3d(lst_of_features)
        out2d = out3d.mean(dim=2)
        feat_target=torch.cat([out2d,uint,iuni],1))
        feat_target=self.compress(feat_target)
        
        preds_target = self.head_target(feat_target)
        return preds_inter, preds_union, preds_diffs, \
               preds_target, preds_target_fz,  \ 
               preds_null_fz, out3d

\end{python}

\section{Modelling Characteristics}

 We use a Linear warm up scheduler for the first epoch, followed by an annealing cosine strategy with warm restarts \citep{loshchilov2016sgdr}. The initial learning rate was set to \texttt{1.e-3}, the half-life was set to 25 epochs and the period to 50 epochs. For training we used the RAdam optimizer \citep{Liu2020On}.

\section{Computational Considerations}
In the extent of this work various computational resources where utilized, that where not available simultaneously. 
The experiments on cifar10 and all UNet-like architectures where run on CSIRO HPC Bracewell (P100 GPUs). The CIFAR10 experiments used a single GPU per run, the UNet ones from 4 to 24 P100 GPUs. The UNet ones on 4 $\times$ P100 GPU for 24 hours.   

The experiments on the ISPRS dataset for the SSG2 architecture where run on Down Under Geoscience (DUG) HPC cluster in Perth, Western Australia. Each run utilized 24 $\times$ A100 (80GB) graphics cards, the training time was 3 days for $N=5$ sequence elements.

The experiments on the Darwin UrbanMonitor dataset  as well as the ISIC2018 where run on AMD MI250 GPUs on the Setonix HPC cluster at Pawsey. The UrbanMonitor dataset 1 tile experiments where run on a single node (4 $\times$ MI250 GPUs), while the experiments with 4 tiles for training on 4 nodes (16 $\times$ MI250). The ISIC2018 SSG2 experiments where run on 2 nodes (8 $\times$ MI250 GPUs).

\begin{table}[ht!]
    \centering
    \fontsize{8}{10}\selectfont
    \begin{tabular}{lccc}
        \toprule
        & UNet & \texttt{SSG2}(N=4)  & \texttt{SSG2}(N=2) \\
        \midrule
        \textbf{Total Parameters (M)} & 72.41 & 75.55 & 75.55 \\
        \textbf{Total Mult-Adds (G)} & 64.95 & 879.04 & 462.27 \\
        \textbf{Input Size (MB)} & 1.31 & 6.55 & 3.93 \\
        \textbf{Fwd/Bwd Pass Size (MB)} & 2665.50 & 18391.06 & 12149.92 \\
        \textbf{Params Size (MB)} & 289.65 & 302.19 & 302.19 \\
        \textbf{Estimated Total Size (MB)} & 2956.46 & 18699.80 & 12456.04 \\
        \bottomrule
    \end{tabular}
    \caption{Comparison of UNet-like model and SSG2 for the case of \texttt{PTA-T-ViT} feature extraction unit.}
    \label{tab:model_comparison}
\end{table}

In Table \ref{tab:model_comparison}, we present a summary of the UNet-like and SSG2 models, both of which employ the \texttt{PTA-T-ViT} as their base feature extractor. The reported parameters and memory footprints are based on an input image size of $(1,5,256,256)$ and include 4 and 2 support images of identical spatial dimensions for the SSG2 model, respectively. Notably, the SSG2 model demonstrates a substantially higher memory footprint  approximately 7-fold and 5-fold greater for sequence lengths$ N=4$ and $N=2$, respectively, compared to the UNet-like model.

As for throughput, we report the following inference times measured on an NVIDIA RTX A3000 GPU. For a single input image, the forward pass for the UNet-like model took 0.0311 seconds. In contrast, the SSG2 model required 0.2176 seconds for $N=2$ sequence elements, making it roughly 7 times slower. When the sequence length was increased to $N=4$, the forward pass took 0.5449 seconds, which is approximately 18 times slower than its UNet-like counterpart.

\end{document}